\pdfoutput=1

\documentclass[10pt,twocolumn,letterpaper]{article}

\usepackage[pagenumbers]{cvpr}

\DeclareMathOperator*{\argmax}{arg\,max}

\usepackage{multirow}\usepackage{colortbl}\newcommand{\grp}[1]{\textcolor{gray}{\emph{#1}}}\newcommand{\sbest}[1]{\underline{#1}}
\newcommand{\bench}{StreamFun3D\xspace}
\newcommand{\ours}{StreamAff\xspace}

\newcommand{\hardPix}{200}

\usepackage{placeins}
\usepackage{alphalph}

\definecolor{cvprblue}{rgb}{0.21,0.49,0.74}
\usepackage[pagebackref,breaklinks,colorlinks,allcolors=cvprblue]{hyperref}

\def\confName{CVPR}
\def\confYear{2027}

\title{When to Respond, Where to Ground: Streaming 3D Affordance Grounding}

\author{
Wenting Luan$^{1}$\quad
Weilong Yan$^{2,\dagger}$\quad
Siqi Lu$^{1}$\quad
Ming Li$^{3}$\\
Beibei Lin$^{2}$\quad
Yongbin Zheng$^{1,*}$\quad
Robby T. Tan$^{2}$\\[2mm]
{\normalsize $^{1}$National University of Defense Technology\quad
$^{2}$National University of Singapore}\\
{\normalsize $^{3}$The Chinese University of Hong Kong, Shenzhen}\\[1mm]
{\small $^{*}$Corresponding author\quad $^{\dagger}$Project leader}
}

\begin{document}
\twocolumn[{
  \renewcommand{\twocolumn}[1][]{#1}
  \maketitle
  \begin{minipage}{\textwidth}
    \centering
    \includegraphics[width=\linewidth]{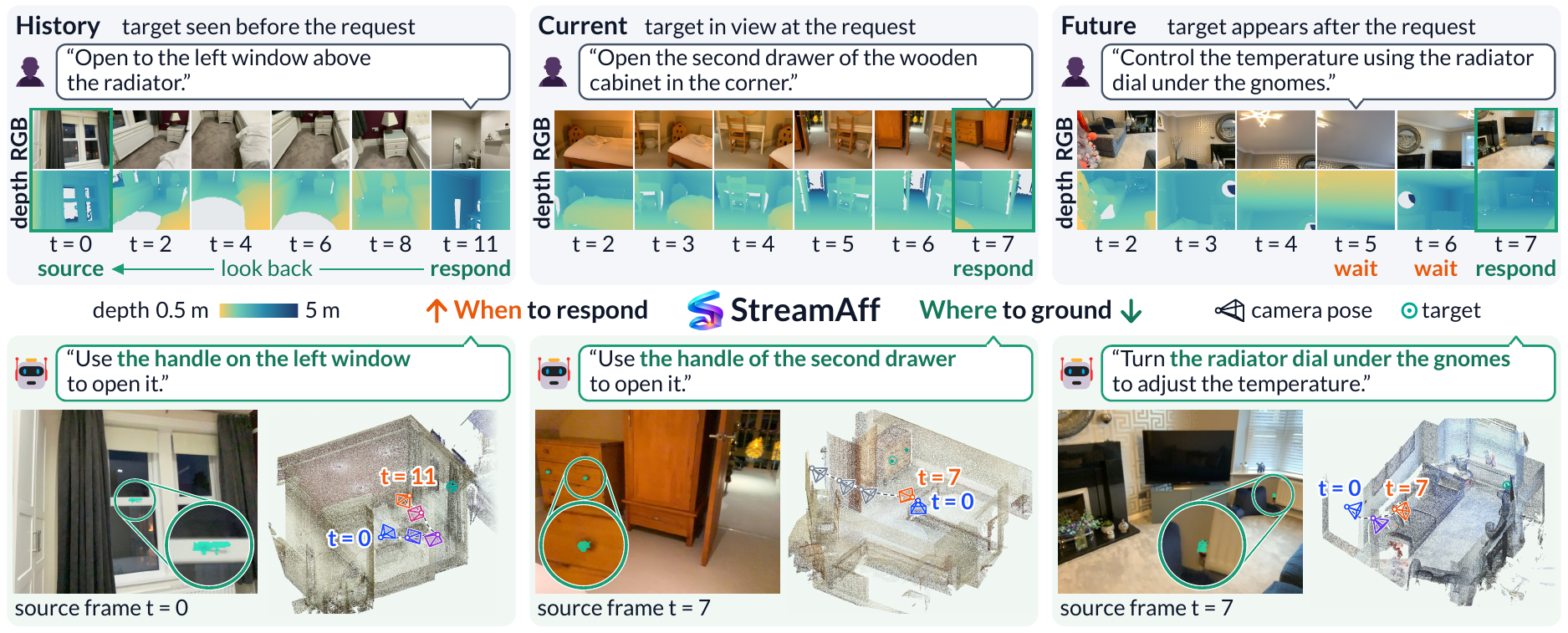}
    \vspace{-8mm}
    \captionsetup{hypcap=false}
    \captionof{figure}{\textbf{When to respond, where to ground.}
    \textbf{History:} a request arrives at $t=11$ and \ours responds using evidence observed at $t=0$.
    \textbf{Current:} a request at $t=7$ is grounded in the same frame.
    \textbf{Future:} a request arrives at $t=5$; the agent waits until $t=7$ to respond.
    Each response identifies the functional part in language and predicts its 3D mask. RGB-D observations, camera trajectories, and mask close-ups illustrate the three cases.}
    \label{fig:teaser}
  \end{minipage}
  \par\vspace{1em}
}]
\begin{abstract}
Given a continuous RGB-D stream and a language request, streaming 3D affordance grounding requires an agent to decide when to respond and where to locate the requested functional part using only observations available so far.
Existing 3D affordance grounding methods typically assume complete scene observations, while streaming vision-language models lack fine-grained 3D functional grounding.
Our goal is to jointly determine response timing and functional-part localization during continuous scene observation.
We propose \textbf{StreamAff}, an evidence-guided response strategy that makes both decisions from the grounding model's own visual evidence.
The model scores the visual patches of the frames it reads: their combined support, weighed against a wait option, decides when to respond, and the highest-scoring patch selects the frame to ground in.
A conditional look-back policy lets the model answer from an earlier observation at the moment of the request or wait for new evidence, and zoom-and-refine grounding segments the part named in the answer on the 3D points of the selected frame.
We further introduce \textbf{StreamFun3D}, a benchmark derived from real-world 3D scenes that evaluates response timing and grounding across four evidence settings:
Current, History, Future, and Unobserved, together with StreamIoU, a metric that credits a mask only when the response is timely and grounded in a valid source frame.
We adapt offline 3D affordance grounding methods and build external-decider baselines that pair vision-language models with our mask decoder. StreamAff achieves 30.14\% StreamIoU on 15-second streams, a 40.6\% relative improvement over the strongest of these baselines.
\end{abstract}

\setlength{\textfloatsep}{10pt plus 2pt minus 2pt}
\setlength{\dbltextfloatsep}{10pt plus 2pt minus 2pt}
\setlength{\floatsep}{8pt plus 2pt minus 2pt}
\setlength{\dblfloatsep}{8pt plus 2pt minus 2pt}
\setlength{\abovecaptionskip}{5pt}
\setlength{\belowcaptionskip}{2pt}
\vspace{-28pt}
\section{Introduction}
\label{sec:intro}

Grounding language instructions to functional parts in 3D
scenes is important for robots and assistive agents
\cite{delitzas2024scenefun3d,corsetti2025fun3du,
wang2025affordbot}.
For example, a user may ask a robot to ``open the window
above the sink'' while the robot is observing a room.
The robot must identify the correct window and locate its
handle in 3D.
However, the handle may not be visible when
the request arrives.
It may have been seen earlier or may appear later.
Responding too early may lead to an
incorrect part, while waiting unnecessarily delays the
response.
The challenge is not only to locate
the requested part, but also to determine when sufficient
visual evidence is available.

Existing 3D affordance grounding methods, such as Fun3DU
and T-FunS3D, typically operate on complete scene
representations or previously captured observations
\cite{corsetti2025fun3du,feng2026tfuns3d}.
Recent methods such as UniFunc3D~\cite{lin2026unifunc3d} further improve affordance grounding by actively selecting informative
frames and regions.
Although these methods can identify functional parts
from language, they assume that the required observations
are already available.
They are not able to decide
whether the requested part has been observed or whether
the system should wait for future observations.
Applying these methods independently to incoming frames
would also fail to fully use evidence from the past.

A request can specify a story to generate~\cite{shi2026one}, a subject to discuss in a video~\cite{pvchat}, or a scientific process to explain~\cite{xu2026scieducator}, while a streaming vision-language model must also decide when the incoming observations support a response
\cite{chen2024videollmonline,wang2024mmduet,
azad2026streamready,zeng2026onestreamer}.
Stream3D-VLM \cite{yu2026stream3dvlm} extends this capability to online 3D
spatial understanding, while
TempoGround \cite{ding2026tempoground} models object correspondence and presence
during streaming grounding.
However, these methods do not jointly address response timing and fine-grained functional-part grounding.
One possible solution is to combine a separate response decision model with an existing grounding model.
In such a cascade, however, the decision model judges whether the observed evidence is sufficient using its own representation, rather than the evidence from which the grounding model predicts the 3D mask.
The decision to respond is therefore not conditioned on whether the requested part can actually be grounded, which matters when similar objects or functional parts appear at different times.
We evaluate such cascades in \cref{sec:exp} by pairing vision-language deciders with our mask decoder.

In this paper, we study \emph{streaming 3D affordance grounding}: given an RGB-D stream and a request introduced during the stream, an agent must respond as soon as the observed frames provide valid evidence for the requested functional part, returning an answer and a 3D mask on a valid source frame, and must keep waiting otherwise.
To evaluate this task, we build \textbf{\bench} from SceneFun3D, with supplementary data from ScanNet++~\cite{delitzas2024scenefun3d,yeshwanth2023scannetpp}.
Frame-level evidence annotations, derived from annotated 3D parts, camera poses, and depth, specify when a response is supported and which frames are valid sources.
\bench covers four evidence settings, Current, History, Future, and Unobserved, and thus measures whether a model can use past evidence, wait for future evidence, and withhold a response when the requested part is never observed.
Its metric, StreamIoU, credits a mask only when the response is timely and grounded in a valid source frame.

To address challenges in this task, we propose \textbf{\ours}, an evidence-guided response strategy that makes the response decision from the grounding model's own visual evidence rather than from a separate decision model.
Building on a streaming baseline derived from Ground3D-LMM~\cite{ground3dlmm}, \ours scores the visual patches of the frames it reads; their combined support, weighed against a wait option and trained with wait supervision, decides when to respond, and the highest-scoring patch selects the source frame.
A \emph{conditional look-back policy} reads the current frame first, looks back over the stored frames only when it offers no evidence, and afterwards reads only new frames, so that the model can answer from an earlier observation at the request or wait for new evidence.
\emph{Zoom-and-refine grounding} then zooms into the regions that match the part named in the answer and refines the mask voxel by voxel.

Our main contributions are summarized as follows:
\begin{itemize}
  \item We introduce \emph{streaming 3D affordance grounding}, which requires determining both when to respond and where to locate a requested functional part as observations arrive.
  \item We construct \textbf{\bench}, a benchmark built on real-world 3D scene captures with frame-level annotations and 3D functional-part masks across four evidence settings, and propose StreamIoU, a metric that jointly scores response timing, source validity, and mask quality.
  \item We propose \textbf{\ours}, an evidence-guided response strategy in which the combined support of visual patches decides when to respond and the highest-scoring patch selects the source frame, complemented by a conditional look-back policy and zoom-and-refine grounding. On 15-second streams of \bench, \ours achieves 30.14\% StreamIoU, 40.6\% higher (relative) than the strongest constructed baseline.
\end{itemize}

\vspace{-6pt}
\section{Related work}
\label{sec:related}

\paragraph{Functional and affordance grounding in 3D.}
SceneFun3D~\cite{delitzas2024scenefun3d} annotates functional parts in real indoor scenes together with the task descriptions that refer to them, and has become the standard benchmark for task-driven part grounding.
Most methods lift 2D reasoning to 3D: Fun3DU~\cite{corsetti2025fun3du} parses the task with a language model, segments the part in selected views, and fuses the views by voting, while T-FunS3D~\cite{feng2026tfuns3d} builds a hierarchical open-vocabulary representation.
Other approaches use functional scene graphs~\cite{zhang2025openfungraph,hu2026hierfungraph}, memory of previously seen scenes~\cite{wang2026affordmem}, proposal selection learned with reinforcement learning~\cite{lin2026thinkafford}, or large multimodal models that reason over the whole scene or video~\cite{lin2026unifunc3d,wang2025affordbot,huang2024reason3d}.
2D affordance models~\cite{yuan2024robopoint,ma2024glover,ma2025gloverpp,tang2025uad,wu2025ragnet} predict where to act within a single image.
All these methods observe the complete scan or video before answering, so they never decide \emph{when} to respond and always produce an answer.
Like completing a partial shape~\cite{Yan_2026_CVPR} or synthesizing a scene from a coarse proxy~\cite{Xu2026Proxy2WorldLT}, localizing an \emph{invisible} functional object in 2D with SceneFunRI~\cite{chen2026scenefunri} requires inferring what is missing, whereas our agent waits until the part becomes visible and withholds a response if it never does.

\paragraph{Streaming video and 3D perception.}
Streaming video-language models decide when to speak and retain long-range evidence in compact memories or key-value caches~\cite{chen2024videollmonline,wang2024mmduet,qian2025dispider,azad2026streamready,zhang2024flashvstream,di2025rekv,kim2025infinipotv,yang2025streammem}, and some respond proactively once the evidence appears~\cite{ma2026responseg1,zeng2026onestreamer}; their responses, however, are text only.
Visual perception methods detect objects in real time in cluttered traffic scenes~\cite{li2026fad} or with energy-efficient spiking networks~\cite{li2025msd}, correct rolling-shutter distortion in a single image~\cite{Yan_2023_ICCV}, estimate depth from video~\cite{Yan_2025_CVPR}, or describe motion in dynamic point clouds~\cite{zhang20264dpc}, while streaming 3D reconstruction~\cite{wang2025spann3r,wang2025cut3r,zhuo2026streamvggt,murai2025mast3rslam} and online 3D mapping~\cite{xu2025embodiedsam,gu2024conceptgraphs,werby2024hovsg} update a scene representation as frames arrive without deciding when to answer a request.
Closest to ours, Stream3D-VLM~\cite{yu2026stream3dvlm} emits a continue or stop token per frame, TempoGround~\cite{ding2026tempoground} grounds objects with state tokens that include a null state, and EgoRecall~\cite{tam2026egorecall} and ESOM~\cite{manigrasso2026esom} localize previously seen objects in egocentric streams.
These methods target whole objects and respond at the request or within the current view, whereas \ours grounds small functional parts as 3D masks on a source frame that may lie in the past, arrive later, or never appear.

\FloatBarrier
\setlength{\abovedisplayskip}{5pt plus 2pt minus 2pt}
\setlength{\belowdisplayskip}{5pt plus 2pt minus 2pt}
\setlength{\abovedisplayshortskip}{2pt plus 2pt}
\setlength{\belowdisplayshortskip}{3pt plus 2pt minus 1pt}
\section{Benchmark and Evaluation}
\label{sec:bench1}
\label{sec:task}
\vspace{-1mm}

To evaluate when to respond and where to ground, we introduce \bench, a benchmark for streaming 3D affordance grounding with frame-level evidence annotations and 3D functional-part masks.
Each sample pairs an RGB-D stream with a request introduced during the stream; the evidence labels determine whether and when a response is supported, and answerable requests come with masks on the 3D points of the valid source frames and a grounded reference answer (\cref{fig:bench}).

\subsection{Task Definition}
\label{sec:task:def}
\vspace{-1mm}
An agent observes an RGB-D stream $o_1,\ldots,o_T$, one frame per step, and receives a natural-language request at step $q$.
At each step $t\ge q$, using only the request and observations $o_1,\ldots,o_t$, it either waits or responds.
Each request permits at most one response, consisting of a natural-language answer, a nonempty set of observed source frames $\hat{\mathcal{S}}\subseteq\{1,\ldots,t\}$, and a mask of the functional part needed to fulfill the request on the 3D points of each source frame.

\begin{figure*}[t]
  \centering
  \includegraphics[width=\textwidth]{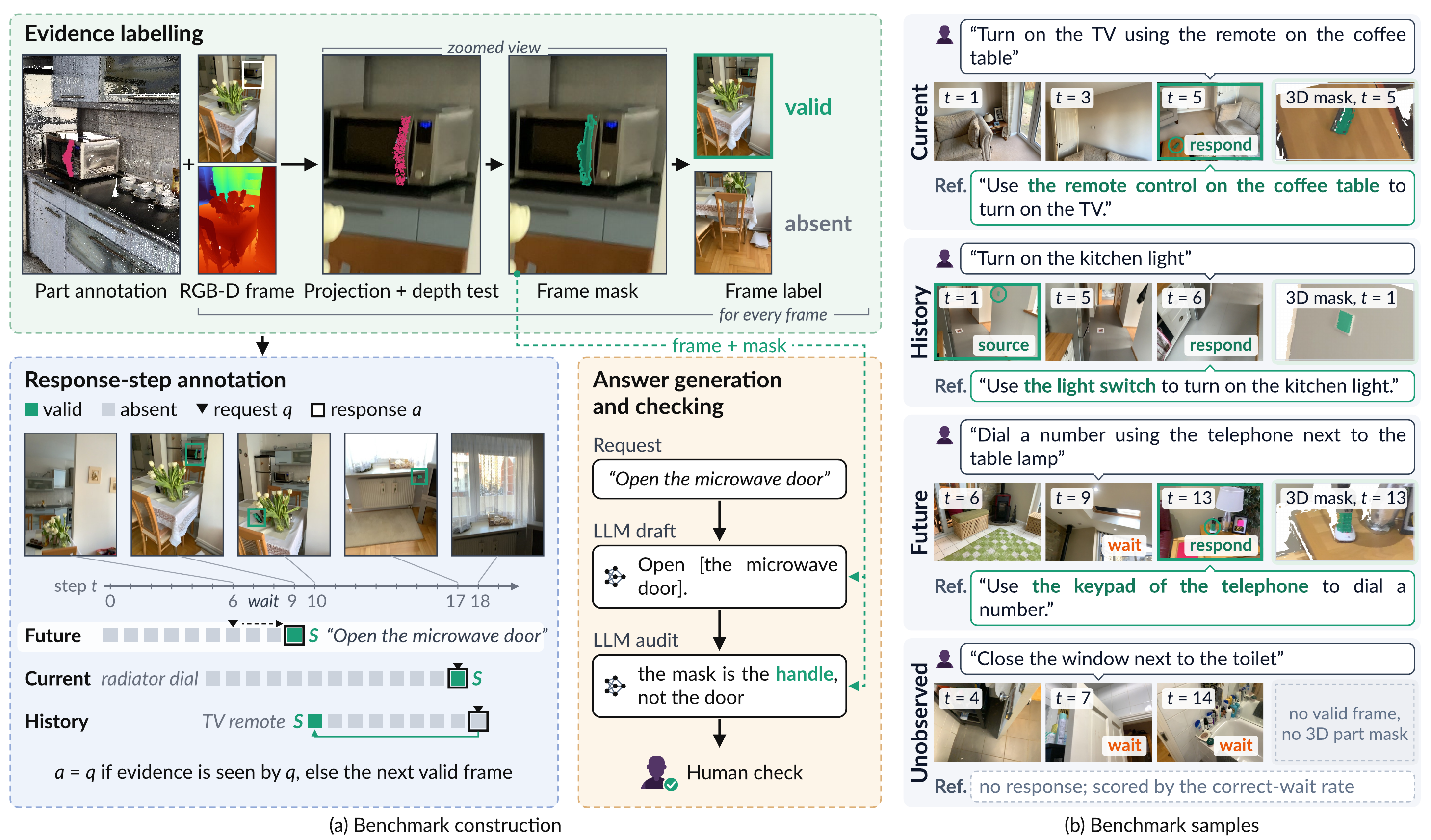}
  \caption{\textbf{Construction and evidence settings of \bench.}
  (a)~Construction of a grounded streaming sample. (b)~Examples of the four evidence settings. Green marks valid source frames and target parts.}
  \label{fig:bench}
  \label{fig:bench_examples}
  \vspace{-1.5mm}
\end{figure*}

For answerable requests, the required response step $a$ is the earliest step at or after $q$ by which a valid source frame has been observed.
The availability of valid evidence defines four settings (\cref{fig:bench}(b)):
(1) \textbf{Current}: the frame at $q$ is a valid source, so $a=q$;
(2) \textbf{History}: the frame at $q$ is not a valid source, but at least one earlier frame is, so $a=q$ and the response must use past evidence;
(3) \textbf{Future}: the first valid source arrives at $a>q$;
and (4) \textbf{Unobserved}: no frame in the stream is a valid source, so the agent must wait throughout the stream.

\subsection{StreamFun3D Construction}
\label{sec:bench2}
\label{sec:bench}
\vspace{-1mm}
We build \bench on SceneFun3D~\cite{delitzas2024scenefun3d} and ScanNet++~\cite{yeshwanth2023scannetpp}, using Articulate3D~\cite{halacheva2024articulate3d} part annotations and OpenTarget~\cite{huang2025openground} descriptions for the latter.
SceneFun3D provides 3{,}200 training clips and 650 evaluation requests for each stream length (15\,s and 30\,s); ScanNet++ adds 892 training clips and 294 evaluation requests.
Streams are sampled at 1\,fps, with disjoint training and evaluation scenes (\cref{fig:bench}(a)).

\vspace{-12pt}
\paragraph{Evidence labelling.}
We project part annotations into each frame using camera poses and depth to obtain masks on its 3D points.
Frames are \emph{valid} when visibility, size, and geometric consistency checks pass, \emph{absent} when the part is verified to be out of view, and unlabelled otherwise.
Camera poses are used for annotation only.

\vspace{-12pt}
\paragraph{Response-step annotation.}
The evidence timeline instantiates the four settings in \cref{sec:task:def} and determines the required response step $a$ and valid source frames $\mathcal{S}$ observed by $a$, without manual timing labels.
Unobserved requests require all frames to be absent; requests whose settings depend on unlabelled frames are discarded.

\vspace{-12pt}
\paragraph{QA annotation.}
Reference answers name the functional part required by each request.
SceneFun3D retains its task descriptions, with answers drafted and verified by vision-language models; ScanNet++ request--answer pairs are rewritten from part descriptions and checked for target consistency and disambiguation.
All evaluation samples undergo a final manual review.

\vspace{-12pt}
\paragraph{Sampling and difficulty.}
Each SceneFun3D evaluation set contains 150 requests per answerable setting and 200 Unobserved requests (\cref{tab:stats}).
\begin{table}[t]
\caption{\textbf{\bench evaluation sets.} Ages and waiting times are in steps (seconds).}
\label{tab:stats}
\centering
\footnotesize
\setlength{\tabcolsep}{3.5pt}
\renewcommand{\arraystretch}{0.92}
\begin{tabular}{@{}l cc@{}}
\toprule
 & 15\,s streams & 30\,s streams \\
\midrule
Frames per stream (1\,fps)               & 15 & 30 \\
Requests per answerable setting         & 150 & 150 \\
Unobserved requests                            & 200 & 200 \\
Answerable requests, Easy / Hard         & 225 / 225 & 224 / 226 \\
Scenes / unique frames                   & 30 / 8{,}346 & 28 / 14{,}584 \\
Request step, mean (range)               & 8.0 (5--12) & 13.1 (5--27) \\
History evidence age, mean (range)          & 4.9 (1--12) & 7.6 (1--27) \\
Future waiting time, mean (range)          & 3.2 (2--8) & 5.0 (2--19) \\
Valid sources per answerable request     & 1.6 & 1.8 \\
Words per request                        & 9.0 & 9.0 \\
\bottomrule
\end{tabular}
\vspace{-1.5mm}
\end{table}

Answerable requests are Hard if the minimum projected part area across their valid source frames is below \hardPix{} pixels or instance disambiguation is required, and Easy otherwise.
Projection thresholds, temporal sampling constraints, and QA procedures are detailed in the supplementary material.

\subsection{Evaluating Timely Grounding}
\label{sec:bench3}
\label{sec:metric}
\vspace{-1mm}
Mask accuracy alone does not capture whether an agent responds when the observed evidence supports it.
We introduce \textbf{StreamIoU}, which credits a mask only when the response is timely and its source is valid.
For the $N$ answerable requests, with required step $a_n$, valid sources $\mathcal{S}_n$, response step $t_n$, and reported sources $\hat{\mathcal{S}}_n$,
\begin{equation}
  \mathrm{StreamIoU}
  = \frac{1}{N}\sum_{n=1}^{N}
    \underbrace{\mathbf{1}[t_n=a_n]}_{\mathrm{T}_n}\;
    \underbrace{\frac{|\hat{\mathcal{S}}_n \cap \mathcal{S}_n|}{|\hat{\mathcal{S}}_n|}}_{\mathrm{E}_n}\;
    \mathrm{IoU}_n,
  \label{eq:smiou}
\end{equation}
where $\mathrm{IoU}_n$ is the mean mask IoU over the frames in $\hat{\mathcal{S}}_n \cap \mathcal{S}_n$, or zero if the intersection is empty.
Early, late, and missing responses thus score zero, and an invalid source lowers the score even when a valid one is also reported.

\vspace{-3mm}
\paragraph{Delay tolerance.}
Because the strict gate also zeroes a response that is one step late, we additionally report \textbf{StreamIoU@$k$}, which replaces $\mathrm{T}_n$ with $\mathbf{1}[a_n \le t_n \le a_n+k]$ and evaluates sources and masks against the valid frames observed by the actual response step.
Early responses still receive no credit, and $k=0$ recovers the strict metric.

\vspace{-3mm}
\paragraph{Correct waiting and diagnostics.}
For Unobserved requests, we report the \textbf{correct-wait rate}, the fraction on which the agent withholds a response until the end of the stream.
\emph{mIoU} measures mask quality without the timing requirement: a streaming agent is triggered at $a_n$, selects its sources, and predicts their masks, while an offline method is evaluated on the valid source frames of the complete stream.
We rank methods by strict StreamIoU and report it together with the correct-wait rate.

\FloatBarrier
\begin{figure*}[t]
  \centering
  \includegraphics[width=\textwidth]{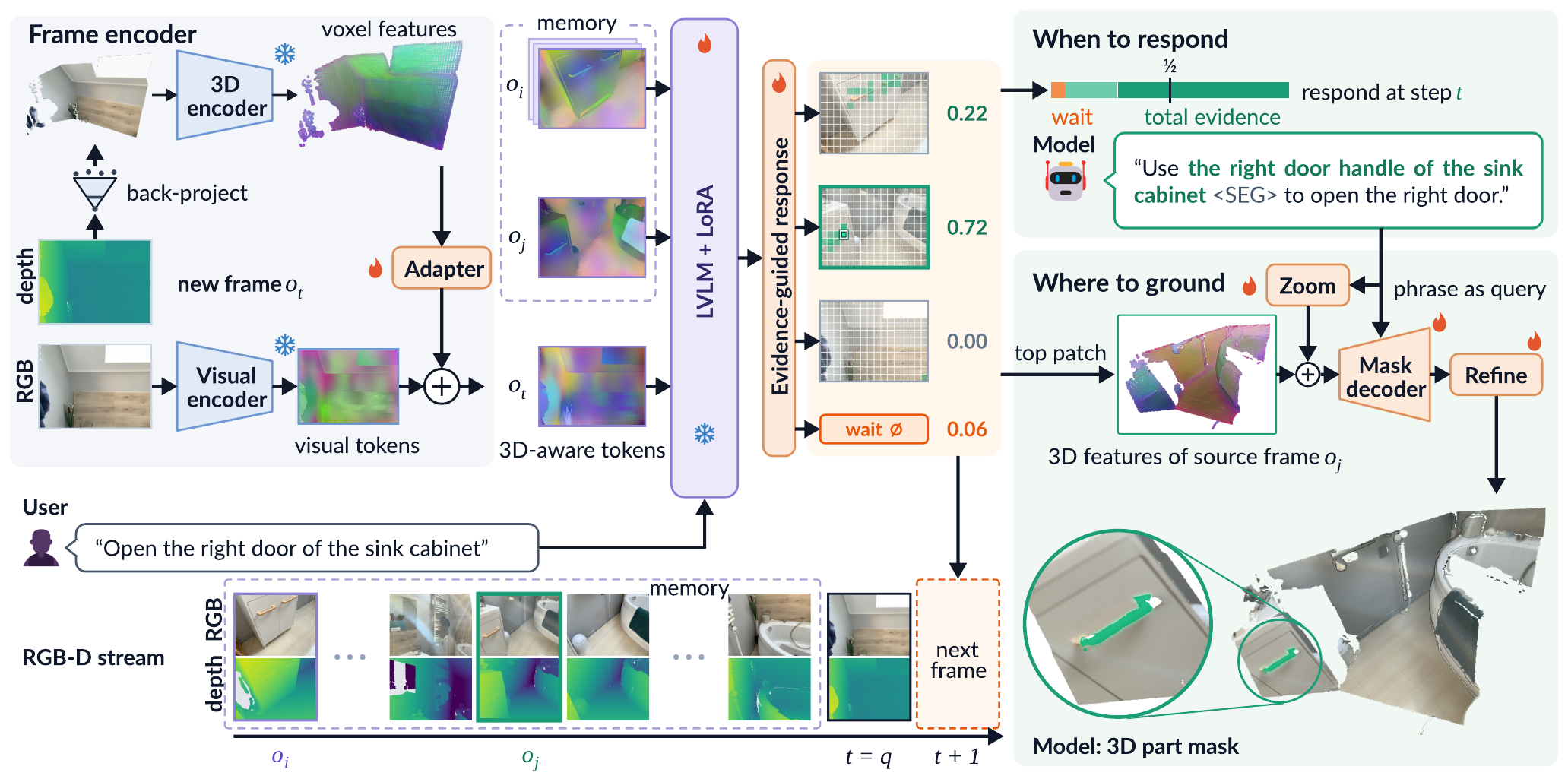}
  \caption{\textbf{Overview of \ours.}
  RGB-D observations are stored as 3D-aware visual tokens.
\emph{When to respond:} at request step $q=11$, the current frame yields \textsc{Wait} ($p_{\varnothing}=0.99$); looking back over all 12 observed frames lowers $p_{\varnothing}$ to $0.06$, triggering a response at the same step. Scores denote probability masses in this look-back read.
\emph{Where to ground:} the outlined top-scoring patch selects an earlier frame ($i^\star=4$). The model answers from the look-back read and uses its part phrase to segment that frame's 3D points. Zoom re-encodes phrase-selected windows at higher resolution and adds their features to the frame's 3D features; Refine corrects the decoder mask voxel-wise.}
  \label{fig:method}
  \vspace{-1.5mm}
\end{figure*}

\section{Proposed Method}
\label{sec:method}
\vspace{-1mm}

The four evidence settings in \cref{sec:task:def} require an agent to decide, at every step after a request, whether the frames observed so far support a response and, if so, which frame to ground it on; StreamIoU (\cref{sec:metric}) credits a response only when both decisions are correct.

\subsection{Framework Overview}
\label{sec:method:overview}
\label{sec:pointer}
\vspace{-1mm}
To make both decisions, we introduce \ours, an evidence-guided response strategy that learns when to respond and where to ground from visual evidence for the requested functional part (\cref{fig:method}).
Building on Ground3D-LMM~\cite{ground3dlmm}, we extend 3D grounding to a continuous stream of RGB-D observations.
Its evidence-guided response, trained with wait supervision and read through a conditional look-back policy, determines when to respond and selects the source frame (\cref{sec:method:when}), while zoom-and-refine grounding localizes the requested functional part on the 3D points of that frame (\cref{sec:method:where}).

\vspace{-5mm}
\paragraph{Streaming baseline.}
\label{sec:method:read}
\label{sec:read}
We first construct a streaming baseline that responds at every request step.
As the stream unfolds, the baseline encodes each RGB-D frame once and stores it as $N$ 3D-aware visual tokens representing image patches.
When a request arrives at step $q$, the language model reads the frames observed so far, denoted by $\mathcal{R}$, in temporal order, followed by the request.
A pointing head assigns each patch a grounding confidence $p_{ij}$, the probability that patch $j$ of frame $i$ shows the requested part, and selects the source frame as $i^\star = \arg\max_{i\in\mathcal{R}}\max_j p_{ij}$.
The model then generates its answer and, following Ground3D-LMM\cite{ground3dlmm}, uses the mean hidden state $\mathbf{z}_{\mathrm{part}}$ of the part phrase as the grounding query:
\begin{equation}
    \widehat{M}_{i^\star} = \mathcal{G}(\mathbf{z}_{\mathrm{part}}, o_{i^\star}),
    \label{eq:mask}
\end{equation}
where $\mathcal{G}$ predicts a mask on the 3D points of the selected observation $o_{i^\star}$ using its voxel features.
The pointing head is trained to assign probability mass to $\mathcal{P}$, the set of patches in $\mathcal{R}$ that overlap projected target-part pixels in valid source frames. This baseline selects its own source frame, but it always responds at the request step and decodes the mask in a single pass from whole-frame features.

\subsection{Evidence-Guided Response}
\label{sec:method:when}
\label{sec:method:wop}
\label{sec:method:train}
\vspace{-1mm}
To let the baseline wait until it is confident that the requested part has been observed, we add a wait option $\varnothing$ to the pointing head.
For each read $\mathcal{R}$, we normalize the wait score jointly with all patch scores, so that the combined support of its patches, the grounding confidence of the read, is:
\begin{equation}
    p_{\mathrm{resp}} = \sum_{i\in\mathcal{R}}\sum_{j=1}^{N} p_{ij} = 1-p_{\varnothing}.
    \label{eq:scores}
\end{equation}
The agent responds when this confidence exceeds $p_{\varnothing}$ and waits otherwise.
The same confidence thus decides both when to respond and where to ground: the agent can respond only by placing probability on observed patches, so every response comes with the frame that supports it.

\vspace{-5mm}
\paragraph{Wait supervision.}
We derive read-level targets from the frame-level evidence labels of \bench (\cref{sec:bench}), without manual timing labels.
Reads containing a valid source frame receive response targets, while reads whose frames are all labelled absent receive wait targets.
Reads satisfying neither condition provide no decision supervision.
The resulting loss is
\begin{equation}
    \mathcal{L}_{\mathrm{wp}} =
    \begin{cases}
        -\log p_{\varnothing}, & \text{wait target},\\[2pt]
        -\log \displaystyle\sum_{(i,j)\in\mathcal{P}} p_{ij}, & \text{response target}.
    \end{cases}
    \label{eq:point}
\end{equation}
\vspace{-5mm}
\paragraph{Conditional look-back policy.}
At the request step, training samples either the current frame or all observed frames, with targets determined by the evidence in that read.
Thus, a current-frame read can be supervised to wait while a look-back read at the same step is supervised to respond.
Because a current-frame read can now return \textsc{Wait}, the agent first reads the current frame at step $q$ and, if the head returns \textsc{Wait}, looks back once over the stored frames.
If both reads return \textsc{Wait}, it continues observing and reads each newly arrived frame, $\mathcal{R}=\{t\}$, until it responds or the stream ends.

\subsection{Zoom-and-Refine Grounding}
\label{sec:method:where}
\label{sec:method:respond}
\vspace{-1mm}
Within the selected source frame, functional parts such as handles and switches often occupy small image regions.
A single decoding pass using whole-frame features can produce coarse or incomplete masks.
We address this by using the part phrase to guide local zooming, followed by voxel-level mask refinement.

\vspace{-10pt}
\paragraph{Zoom.}
The zoom-in branch uses the part query to identify regions for closer inspection.
At each image location $u$, it computes a similarity score between the part query and the local image feature $\mathbf{d}_u$:
\begin{equation}
    a_u = \tau\,\cos\!\big(\mathbf{W}_q\mathbf{z}_{\mathrm{part}},\,\mathbf{W}_k\mathbf{d}_u\big),
    \label{eq:zoom}
\end{equation}
where $\mathbf{W}_q$ and $\mathbf{W}_k$ project the query and image features into a shared space, and $\tau$ scales the similarity scores.
We score each window on a fixed multi-scale grid using its strongest responses, preserving evidence from small parts that would otherwise be diluted by the background.
The top-scoring fine and coarse windows are cropped and re-encoded at higher resolution.
The resulting crop features are pooled onto the corresponding voxels and added to the voxel features supplied to the mask decoder $\mathcal{G}$, which predicts an initial mask score $m_v$ for each voxel $v$.

\vspace{-1mm}
\paragraph{Refine.}
The initial mask identifies the target region but may have imprecise boundaries.
A voxel refinement head revisits a candidate region $\mathcal{C}$ comprising voxels with non-negligible initial scores and their immediate neighbours.
For each candidate voxel, the head gathers the decoder's mask feature, the initial score, the voxel's colour, and its position relative to the predicted part.
It then exchanges messages with the 26 neighbouring voxels and predicts a residual correction $r_v$:
\begin{equation}
    \widehat{m}_v = m_v + r_v, \qquad v\in\mathcal{C}.
    \label{eq:refine}
\end{equation}
Voxels outside $\mathcal{C}$ are assigned to the background, and thresholding $\widehat{m}_v$ gives the final mask $\widehat{M}_{i^\star}$.
Both stages operate only on the selected source frame, so their computational cost is independent of the number of stored frames.

\vspace{-3mm}
\paragraph{Training.}
All trainable components are jointly optimized with
\begin{equation}
    \mathcal{L}_{\mathrm{total}}
    = d\,\mathcal{L}_{\mathrm{wp}}
    + g\,(\mathcal{L}_{\mathrm{ans}}+\mathcal{L}_{\mathrm{grd}}),
    \label{eq:total_loss}
\end{equation}
where $(d,g)=(1,0)$ for wait targets, $(1,1)$ for response targets, and $(0,1)$ for additional grounding samples.
$\mathcal{L}_{\mathrm{ans}}$ is reference-answer cross-entropy, while $\mathcal{L}_{\mathrm{grd}}$ supervises decoder masks, zoom regions, and refined masks over valid source frames.

\FloatBarrier
\section{Experiments}
\label{sec:exp}
\vspace{-1mm}

\begin{table*}[!t]
\centering
\caption{\textbf{Results on the SceneFun3D evaluation sets of \bench.} Cells report mIoU\,/\,StreamIoU (\%). mIoU ignores timing; StreamIoU (\cref{eq:smiou}) additionally requires the correct response step and a valid source frame. Wait denotes the correct-wait rate (\%) on Unobserved requests. \textbf{Best} and \sbest{second-best} scores are highlighted per metric.}
\label{tab:main}
\footnotesize
\setlength{\tabcolsep}{4pt}
\renewcommand{\arraystretch}{0.94}
\resizebox{0.9\linewidth}{!}{%
\begin{tabular}{@{}l cccccc@{\hskip 10pt}c@{}}
\toprule
 & \multicolumn{6}{c}{mIoU $\uparrow$\,/\,StreamIoU $\uparrow$} & Wait $\uparrow$ \\
\cmidrule(lr){2-7}
Method & Cur. & Hist. & Fut. & Easy & Hard & \textbf{Avg.} & Unobs. \\
\midrule
\multicolumn{8}{@{}l}{\grp{15\,s streams}} \\
Sa2VA-8B~\cite{yuan2025sa2va} & 0.89\,/\,-- & 2.44\,/\,-- & 0.14\,/\,-- & 2.19\,/\,-- & 0.12\,/\,-- & 1.16\,/\,-- & -- \\
GLOVER++~\cite{ma2025gloverpp} & 3.36\,/\,-- & 3.88\,/\,-- & 4.55\,/\,-- & 4.60\,/\,-- & 3.26\,/\,-- & 3.93\,/\,-- & -- \\
AffordanceVLM~\cite{wu2025ragnet} & 7.70\,/\,-- & 6.55\,/\,-- & 8.54\,/\,-- & 9.56\,/\,-- & 5.63\,/\,-- & 7.60\,/\,-- & -- \\
AffordBot$^\dagger$~\cite{wang2025affordbot} & 13.19\,/\,-- & 14.67\,/\,-- & 14.87\,/\,-- & 10.34\,/\,-- & 18.15\,/\,-- & 14.24\,/\,-- & -- \\
\cmidrule(l{2pt}r{2pt}){1-8}
T-FunS3D~\cite{feng2026tfuns3d} & 13.47\,/\,12.04 & 13.30\,/\,13.31 & 13.08\,/\,0.00 & 15.68\,/\,11.19 & 10.89\,/\,5.70 & 13.29\,/\,8.45 & 0.0 \\
Fun3DU~\cite{corsetti2025fun3du} & 19.30\,/\,14.46 & 18.61\,/\,16.20 & 13.31\,/\,0.00 & 22.76\,/\,14.64 & 11.38\,/\,5.80 & 17.07\,/\,10.22 & 0.0 \\
Streaming baseline (\cref{tab:abl}) & 24.66\,/\,24.66 & 18.85\,/\,18.85 & 27.35\,/\,0.00 & 29.23\,/\,18.34 & 18.01\,/\,10.67 & 23.62\,/\,14.51 & 0.0 \\
Qwen3-VL-8B~\cite{bai2025qwen3vl} decides & 32.63\,/\,22.10 & 34.35\,/\,22.21 & 34.21\,/\,13.45 & 41.28\,/\,23.40 & 26.18\,/\,15.11 & 33.73\,/\,19.25 & \textbf{79.5} \\
Qwen3-VL-4B~\cite{bai2025qwen3vl} decides & \sbest{33.83}\,/\,\sbest{24.90} & \textbf{36.30}\,/\,\sbest{25.44} & \sbest{34.69}\,/\,\sbest{13.95} & \sbest{42.70}\,/\,\sbest{24.95} & \sbest{27.18}\,/\,\sbest{17.90} & \sbest{34.94}\,/\,\sbest{21.43} & 76.0 \\
\cmidrule(l{2pt}r{2pt}){1-8}
\textbf{\ours} & \textbf{37.06}\,/\,\textbf{34.28} & \sbest{34.53}\,/\,\textbf{32.85} & \textbf{40.16}\,/\,\textbf{23.30} & \textbf{44.32}\,/\,\textbf{37.93} & \textbf{30.17}\,/\,\textbf{22.36} & \textbf{37.25}\,/\,\textbf{30.14} & \sbest{76.5} \\
\midrule
\multicolumn{8}{@{}l}{\grp{30\,s streams}} \\
Sa2VA-8B~\cite{yuan2025sa2va} & 1.00\,/\,-- & 2.57\,/\,-- & 0.01\,/\,-- & 1.75\,/\,-- & 0.64\,/\,-- & 1.19\,/\,-- & -- \\
GLOVER++~\cite{ma2025gloverpp} & 3.28\,/\,-- & 2.95\,/\,-- & 3.02\,/\,-- & 4.24\,/\,-- & 1.93\,/\,-- & 3.08\,/\,-- & -- \\
AffordanceVLM~\cite{wu2025ragnet} & 5.53\,/\,-- & 6.18\,/\,-- & 6.21\,/\,-- & 7.84\,/\,-- & 4.12\,/\,-- & 5.97\,/\,-- & -- \\
AffordBot$^\dagger$~\cite{wang2025affordbot} & 15.95\,/\,-- & 15.64\,/\,-- & 14.39\,/\,-- & 11.38\,/\,-- & 19.25\,/\,-- & 15.33\,/\,-- & -- \\
\cmidrule(l{2pt}r{2pt}){1-8}
T-FunS3D~\cite{feng2026tfuns3d} & 12.57\,/\,12.07 & 16.04\,/\,11.21 & 13.05\,/\,0.00 & 14.60\,/\,8.30 & 13.19\,/\,7.22 & 13.89\,/\,7.76 & 0.0 \\
Fun3DU~\cite{corsetti2025fun3du} & 19.39\,/\,14.03 & 18.16\,/\,14.65 & 13.64\,/\,0.00 & 22.74\,/\,12.77 & 11.43\,/\,6.37 & 17.06\,/\,9.56 & 0.0 \\
Streaming baseline (\cref{tab:abl}) & 27.53\,/\,\sbest{27.53} & 20.81\,/\,20.81 & 24.88\,/\,0.00 & 28.93\,/\,19.29 & 19.92\,/\,12.96 & 24.40\,/\,16.11 & 0.0 \\
Qwen3-VL-8B~\cite{bai2025qwen3vl} decides & \sbest{33.60}\,/\,22.80 & \textbf{35.10}\,/\,\sbest{22.92} & \sbest{28.69}\,/\,11.33 & \sbest{36.14}\,/\,\sbest{23.16} & 28.83\,/\,14.91 & \sbest{32.47}\,/\,19.02 & \sbest{70.5} \\
Qwen3-VL-4B~\cite{bai2025qwen3vl} decides & 33.12\,/\,25.35 & 33.50\,/\,20.37 & 28.43\,/\,\sbest{11.70} & 34.49\,/\,20.27 & \sbest{28.91}\,/\,\sbest{18.02} & 31.68\,/\,\sbest{19.14} & 60.5 \\
\cmidrule(l{2pt}r{2pt}){1-8}
\textbf{\ours} & \textbf{37.75}\,/\,\textbf{36.59} & \sbest{33.69}\,/\,\textbf{31.45} & \textbf{36.48}\,/\,\textbf{17.33} & \textbf{40.04}\,/\,\textbf{33.13} & \textbf{31.95}\,/\,\textbf{23.83} & \textbf{35.98}\,/\,\textbf{28.46} & \textbf{72.0} \\
\bottomrule
\end{tabular}
}
\end{table*}
\definecolor{refrow}{gray}{0.92}
\providecommand{\gain}[1]{{\scriptsize(#1)}}
\renewcommand{\topfraction}{0.95}\renewcommand{\dbltopfraction}{0.95}\renewcommand{\bottomfraction}{0.9}\renewcommand{\textfraction}{0.04}
\renewcommand{\floatpagefraction}{0.92}\renewcommand{\dblfloatpagefraction}{0.92}\setcounter{topnumber}{4}\setcounter{bottomnumber}{3}\setcounter{totalnumber}{6}\setcounter{dbltopnumber}{3}

\paragraph{Datasets and metrics.}
We evaluate on the SceneFun3D evaluation sets of \bench with 15\,s and 30\,s streams, and on its ScanNet++ evaluation set.
Following \cref{sec:metric}, we report strict StreamIoU on answerable requests, overall and for the Easy and Hard subsets, together with the correct-wait rate on Unobserved requests.
mIoU separates grounding quality from timing, while StreamIoU@$k$ assesses performance under a delay tolerance of $k$ steps.

\vspace{-5mm}
\paragraph{Baselines.}
As no existing method performs this task, we compare with adapted offline 3D affordance grounding methods and with external-decider baselines.
The offline methods, Fun3DU~\cite{corsetti2025fun3du}, T-FunS3D~\cite{feng2026tfuns3d}, AffordBot~\cite{wang2025affordbot}, AffordanceVLM~\cite{wu2025ragnet}, GLOVER++~\cite{ma2025gloverpp}, and Sa2VA~\cite{yuan2025sa2va}, process the complete stream and are evaluated on annotated valid source frames without a timing requirement, as unranked references; Fun3DU and T-FunS3D are also evaluated on the observed prefix with an immediate response at the request step.
The external deciders, Qwen3-VL-8B and Qwen3-VL-4B~\cite{bai2025qwen3vl}, receive the request and all frames observed so far at each step from the request onward, either wait or select a source frame, and share our mask decoder.
\vspace{-5mm}
\paragraph{Implementation details.}
We initialize \ours from Ground3D-LMM-4B-Joint~\cite{ground3dlmm}, built on Qwen3-VL-4B, adapt the language model with LoRA, and train the task-specific modules while keeping the visual and 3D encoders frozen.
Grounding samples and streaming decision events from SceneFun3D and ScanNet++, covering 15\,s and 30\,s streams at 1\,fps, jointly train part grounding and response timing.
All models, including the ablation variants, are trained for 5{,}040 updates and evaluated at the final checkpoint. Further training details are in the supplementary material.

\subsection{Results}
\label{sec:exp:results}

\paragraph{Streaming 3D affordance grounding.}
On the two SceneFun3D evaluation sets, \ours achieves StreamIoU of 30.14 and 28.46 for 15\,s and 30\,s streams, respectively (\cref{tab:main}).
Using the same mask decoder, \ours outperforms both prompted external deciders, with gains of 10.89 and 9.44 points over Qwen3-VL-8B and 8.72 and 9.32 points over Qwen3-VL-4B on 15\,s and 30\,s streams, respectively.
On 15\,s streams, \ours answers 76.9\% of answerable requests at the required step, compared with 52.9\% and 57.1\% for the 8B and 4B deciders, respectively.
It also correctly withholds responses on 76.5\% and 72.0\% of Unobserved requests, whereas immediate answering fails on all such requests.
Beyond these external-decider baselines, \ours also scores above the request-prefix variants of Fun3DU and T-FunS3D in StreamIoU.
In mIoU, which does not involve response timing, \ours reaches 37.25 and 35.98, above all offline methods, which observe the complete stream but differ from \ours in input and training data; the external deciders, which share our mask decoder, reach 31.68--34.94.

\vspace{-12pt}
\paragraph{Results by evidence setting.}
\ours responds on time to approximately 90\% of both Current and History requests on 15\,s streams, supporting timely responses from current as well as earlier observations.
Future requests expose the need to wait: immediate answering scores zero, while \ours achieves 23.30 StreamIoU on 15\,s streams and 17.33 on 30\,s streams.

\begin{table}[!t]
\centering
\caption{\textbf{Results on ScanNet++.} Each cell reports mIoU\,/\,StreamIoU (\%) on 220 answerable requests, with the same protocol as \cref{tab:main}; Wait is the correct-wait rate (\%) on 74 Unobserved requests from 56 scenes. \textbf{Best} and \sbest{second-best} results are highlighted per metric.}
\label{tab:scan}
\label{tab:scannetpp}
\footnotesize
\setlength{\tabcolsep}{3.5pt}
\renewcommand{\arraystretch}{1.05}
\resizebox{\linewidth}{!}{%
\begin{tabular}{@{}l ccc@{\hskip 8pt}c@{}}
\toprule
 & \multicolumn{3}{c}{mIoU $\uparrow$\,/\,StreamIoU $\uparrow$} & Wait $\uparrow$ \\
\cmidrule(lr){2-4}
Method & 15\,s & 30\,s & \textbf{All} & Unobs. \\
\midrule
Fun3DU~\cite{corsetti2025fun3du} & 20.84\,/\,15.59 & 24.85\,/\,\sbest{18.64} & 22.20\,/\,\sbest{16.63} & 0.0 \\
T-FunS3D~\cite{feng2026tfuns3d} & \sbest{26.15}\,/\,15.05 & 22.38\,/\,13.35 & \sbest{24.87}\,/\,14.47 & 0.0 \\
Qwen3-VL-8B~\cite{bai2025qwen3vl} decides & 23.99\,/\,\sbest{15.76} & 23.29\,/\,14.42 & 23.75\,/\,15.30 & \textbf{79.7} \\
Qwen3-VL-4B~\cite{bai2025qwen3vl} decides & 23.24\,/\,12.77 & \sbest{25.43}\,/\,15.89 & 23.99\,/\,13.83 & 58.1 \\
\midrule
\textbf{\ours} & \textbf{30.96\,/\,24.45} & \textbf{32.39\,/\,23.61} & \textbf{31.45\,/\,24.17} & \sbest{67.6} \\
\bottomrule
\end{tabular}
}
\end{table}
\paragraph{ScanNet++.}
On the ScanNet++ evaluation set, \ours achieves the highest StreamIoU and mIoU of all compared methods (\cref{tab:scan}).

\vspace{-2mm}
\subsection{Qualitative comparisons}
\label{sec:exp:qualitative}

\begin{figure*}[!t]
  \centering
  \includegraphics[width=0.82\textwidth]{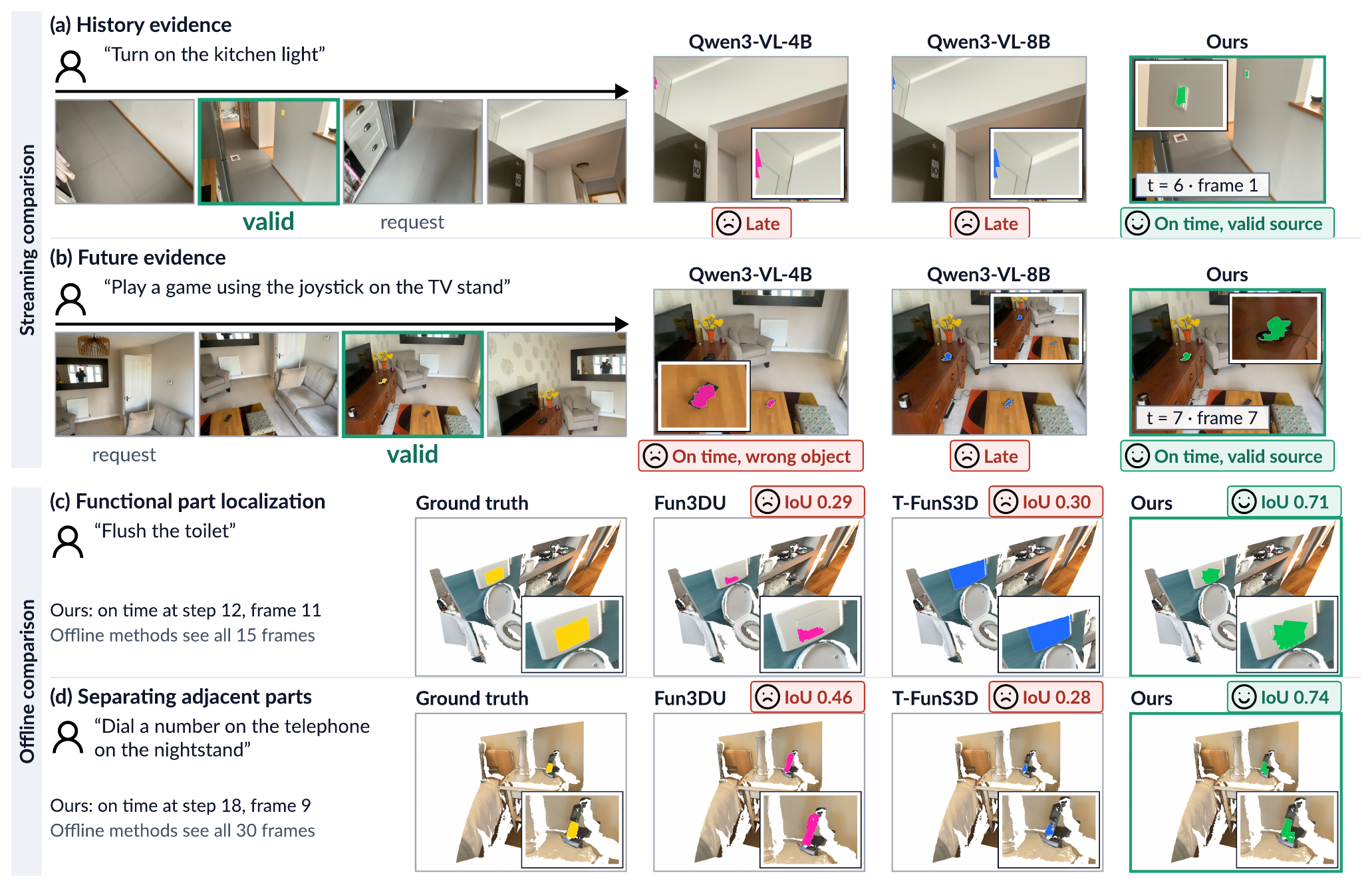}
  \caption{\textbf{Qualitative comparison.}
  \emph{Streaming (a,b)}: all methods see only arrived frames. \ours responds on time by retrieving past evidence (a) or waiting for visibility (b); Qwen3-VL deciders respond late or mask the wrong object.
\emph{Offline (c,d)}: masks are shown on the 3D points of the same valid source frame. \ours better isolates target parts using only frames available at response, while Fun3DU and T-FunS3D use full streams.
Yellow: annotated part; green: \ours; magenta/blue: baselines.}
  \label{fig:qualitative}
\end{figure*}

\Cref{fig:qualitative} shows a streaming and an offline comparison.
In the streaming comparison, \ours grounds a History request on an earlier valid frame at the request step and, for a Future request, waits until the part first comes into view, whereas the external deciders respond late or ground a different object.
In the offline comparison, Fun3DU and T-FunS3D observe the complete stream and locate the requested object, but their masks cover only a fraction of the part or extend onto the rest of the object, whereas the masks of \ours follow the part more closely on the 3D points of the source frame it selected.

\subsection{Ablation}
\label{sec:exp:abl}
We ablate wait supervision, which trains the evidence-guided response, and zoom-and-refine grounding on top of the streaming baseline (\cref{tab:abl}).
All four variants are retrained with the same data and optimization schedule.

\begin{table}[!t]
\centering
\caption{\textbf{Ablation of the two components on SceneFun3D.} StreamIoU, mIoU, and correct-wait rate on Unobserved requests (Wait), all in \%. In parentheses: gain over the streaming baseline, row~(a).}
\label{tab:abl}
\label{tab:ablation}
\footnotesize
\setlength{\tabcolsep}{2.5pt}
\renewcommand{\arraystretch}{1.1}
\resizebox{\linewidth}{!}{%
\begin{tabular}{@{}l@{\hskip 6pt}cc@{\hskip 7pt}lcc@{\hskip 7pt}lcc@{}}
\toprule
 & Wait & Zoom \& & \multicolumn{3}{c}{\textbf{15\,s streams}} & \multicolumn{3}{c}{\textbf{30\,s streams}} \\
\cmidrule(r{6pt}){4-6}\cmidrule(l{1pt}){7-9}
 & sup. & refine & StreamIoU $\uparrow$ & mIoU $\uparrow$ & Wait $\uparrow$ & StreamIoU $\uparrow$ & mIoU $\uparrow$ & Wait $\uparrow$ \\
\midrule
(a) &  &  & 14.51 & 23.62 & 0.0 & 16.11 & 24.40 & 0.0 \\
(b) & \checkmark &  & 18.94\,\gain{+4.43} & 23.96 & \textbf{79.5} & 20.46\,\gain{+4.35} & 24.91 & \sbest{67.5} \\
(c) &  & \checkmark & 23.31\,\gain{+8.80} & \sbest{36.16} & 0.0 & 21.47\,\gain{+5.36} & \sbest{33.29} & 0.0 \\
(d) & \checkmark & \checkmark & \textbf{30.14}\,\gain{+15.63} & \textbf{37.25} & \sbest{76.5} & \textbf{28.46}\,\gain{+12.35} & \textbf{35.98} & \textbf{72.0} \\
\bottomrule
\end{tabular}
}
\vspace{-3mm}
\end{table}
\vspace{-5mm}
\paragraph{Wait supervision.}
Without wait supervision, the model answers every request at the request step: all Future requests are answered prematurely and the correct-wait rate is zero.
Adding it raises StreamIoU from 14.51 to 18.94 on the baseline (rows (a) to (b)) and from 23.31 to 30.14 with zoom-and-refine grounding (rows (c) to (d)) on 15\,s streams, and from 16.11 to 20.46 and from 21.47 to 28.46 on 30\,s streams.
mIoU changes by at most 2.69 points, so the gain comes from when the model responds rather than from better masks.
\vspace{-5mm}
\paragraph{Zoom-and-refine grounding.}
Zoom-and-refine grounding improves the masks themselves: mIoU rises from 23.62 to 36.16 on the baseline (rows (a) to (c)) and from 23.96 to 37.25 with wait supervision (rows (b) to (d)) on 15\,s streams, with similar gains on 30\,s streams.
StreamIoU increases accordingly, by 8.80 and 11.20 points on 15\,s streams.
Together they add 15.63 points over the baseline on 15\,s streams and 12.35 on 30\,s streams; the gains compound because StreamIoU credits a mask only when the response is also timely.

\subsection{Analysis}
\label{sec:exp:analysis}

\paragraph{Delay tolerance.}
Allowing responses up to three steps late raises the StreamIoU of \ours by at most two points, so delays explain only part of the gap to its mIoU, and \ours stays at least 8 points ahead of both external deciders at every tolerance (see the supplementary material).

\paragraph{Latency.}
On a single H200 GPU, median per-frame processing takes 0.25\,s before the request and 0.31\,s afterwards, well within the 1\,s frame interval, and generating the answer with its 3D mask adds 0.76\,s when the agent responds (30 requests from the 15\,s SceneFun3D evaluation set).
The look-back is confined to the request step; later decisions process only the newly arriving frame, and the mask is decoded once on the selected source frame.

\FloatBarrier
\section{Conclusion}
\label{sec:conclusion}
We introduced \textbf{streaming 3D affordance grounding}, where an agent must decide when to respond and which observed frame grounds the requested functional part.
Our benchmark, \textbf{\bench}, provides frame-level evidence annotations, and \textbf{StreamIoU} jointly measures response timing, source validity, and mask quality.
Our method, \textbf{\ours}, lets the combined support of visual patches decide when to respond and the highest-scoring patch select the source frame, with a conditional look-back policy and zoom-and-refine grounding.
It achieves the highest StreamIoU on SceneFun3D and ScanNet++, and the ablations show that wait supervision and zoom-and-refine grounding together raise StreamIoU from 14.51 to 30.14 on 15\,s streams and from 16.11 to 28.46 on the 30\,s streams.
We hope \bench and StreamIoU support future embodied agents that know when to respond and when to wait.
\vspace{-15pt}

{
    \small
    \bibliographystyle{ieeenat_fullname}
    \bibliography{main}
}

\clearpage
\onecolumn
\appendix
\renewcommand\thesubsection{\thesection.\arabic{subsection}}
\crefname{appendix}{Sec.}{Secs.}
\Crefname{appendix}{Section}{Sections}
\crefname{subappendix}{Sec.}{Secs.}
\Crefname{subappendix}{Section}{Sections}
\setcounter{figure}{0}
\setcounter{table}{0}
\renewcommand{\thefigure}{\AlphAlph{\value{figure}}}
\renewcommand{\thetable}{\AlphAlph{\value{table}}}
\renewcommand{\theHfigure}{supp.\arabic{figure}}
\renewcommand{\theHtable}{supp.\arabic{table}}
\makeatletter
\renewenvironment{table}[1][]{\@float{table}[!htb]}{\end@float}
\renewenvironment{table*}[1][]{\@float{table}[!htb]}{\end@float}
\renewenvironment{figure}[1][]{\@float{figure}[!htb]}{\end@float}
\renewenvironment{figure*}[1][]{\@float{figure}[!htb]}{\end@float}
\makeatother
\renewcommand{\topfraction}{.95}
\renewcommand{\bottomfraction}{.9}
\renewcommand{\textfraction}{.05}
\renewcommand{\floatpagefraction}{.85}
\setcounter{topnumber}{4}
\setcounter{bottomnumber}{3}
\setcounter{totalnumber}{6}
\newcommand{\supptab}[1]{Supp/tables/#1}
\graphicspath{{Supp/tables/}{Supp/}}

\begin{center}
{\Large\bf Supplementary Material}
\end{center}
\vspace{2mm}

When a request arrives and the current frame does not show the requested part, a streaming agent has to choose between looking back at what it has seen and waiting for what comes next: a History request must be answered at once from an earlier frame, a Future request only once the part appears.
\cref{sec:supp_lookback} examines how StreamAff makes this choice; \cref{sec:supp_where} turns to what follows a response, the selected source frame and the mask; \cref{sec:supp_benchmark} documents the construction of the benchmark and the data splits; and \cref{sec:supp_impl} gives the implementation details and the baseline protocols.
The additional analyses are run on the two SceneFun3D evaluation sets: two numbers separated by a slash refer to 15\,s / 30\,s streams, and $^*$ marks a paired difference to StreamAff whose 95\% bootstrap interval over requests excludes zero.

\section{Looking Back or Waiting}
\label{sec:supp_lookback}

\begin{figure*}[t]
\centering
\includegraphics[width=\linewidth]{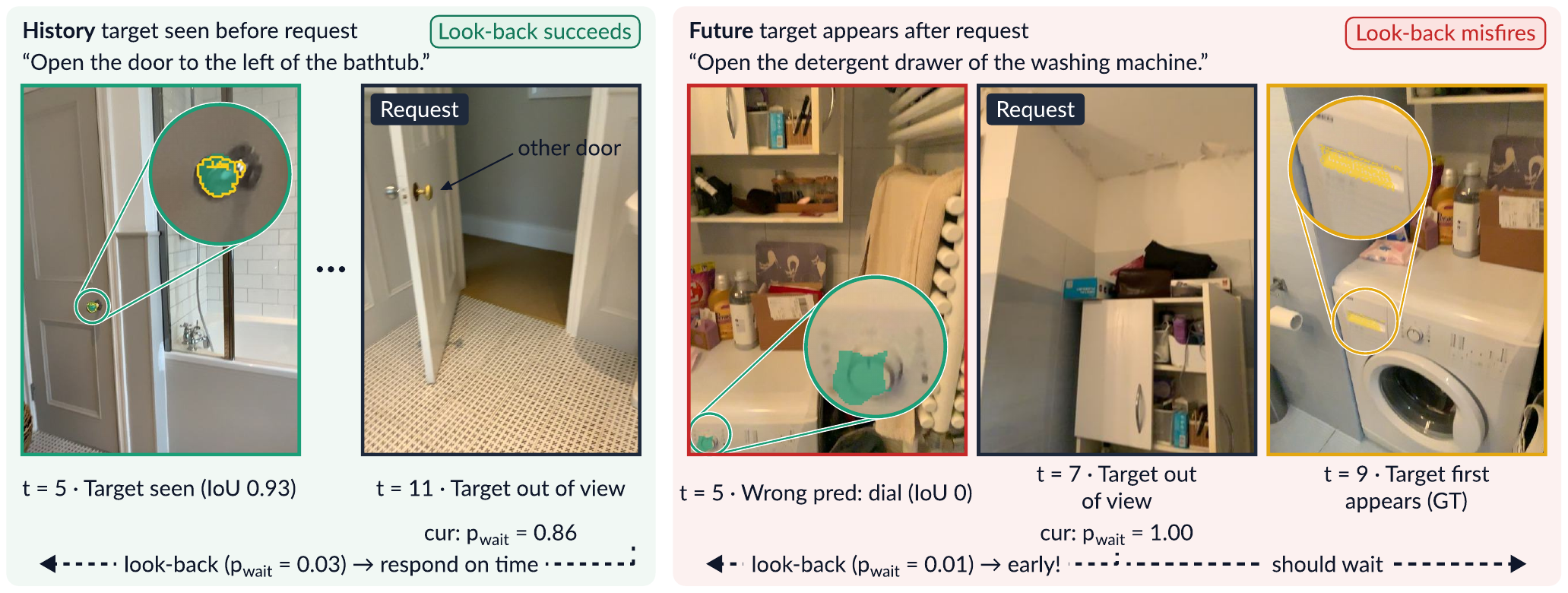}
\caption{\textbf{Two requests that arrive while the requested part is out of view} (SceneFun3D, 15\,s streams). \emph{History:} the request
arrives at $t{=}11$; StreamAff looks back and answers from the frame of $t{=}5$, where its mask covers the door handle (IoU 0.93). The
handles in view at $t{=}11$ belong to another door. \emph{Future:} the request arrives at $t{=}7$ and the drawer first comes into view at
$t{=}9$; StreamAff looks back and answers at $t{=}7$ from the frame of $t{=}5$, where its mask falls on the program dial of the washing
machine (score 0). The insets enlarge the masks. Green: StreamAff mask; yellow: annotated part.}
\label{fig:supp_pair}
\end{figure*}

For the History request in \cref{fig:supp_pair}, looking back is what the task requires: the handle is found in a frame observed six steps earlier, and the response is on time.
For the Future request, the same look-back gives a response two steps before the requested part comes into view.

\paragraph{The look-back.}
Without the look-back, the History StreamIoU on 15\,s streams falls from 32.85 to 0, and the number of early responses to Future requests falls from 33 to 16 (\cref{tab:supp_lookback}).
Removing the look-back also improves correct waiting on Unobserved requests (\cref{tab:supp_lookback}).
At the request step StreamAff reads the current frame first and looks back over the stored frames only if this read returns \textsc{Wait}; a response triggered by the current frame is given without reading the stored frames, and after the request step every new frame is read alone.
For completeness, \cref{tab:supp_lookback} also probes applying the trained model without retraining to a full-history read at every step; because post-request training reads contain only the new frame (\cref{sec:supp_training}), this out-of-distribution policy raises false triggers on longer histories.

\begin{table}[t]
\centering\small
\caption{\textbf{What the look-back contributes and what it costs.} Three read policies on the same model, without retraining
(SceneFun3D 15\,s / 30\,s). Early Future: number of the 150 Future requests that are answered before the required step. Wait: correct-wait
rate on Unobserved requests (\%).}
\label{tab:supp_lookback}
\setlength{\tabcolsep}{5pt}
\begin{tabular}{@{}l cc c c@{}}
\toprule
 & \multicolumn{2}{c}{StreamIoU} & & \\
\cmidrule(lr){2-3}
Frames read & Avg. & History & Early Future & Wait \\
\midrule
Current frame, then each new frame (no look-back) & 19.64$^*$ / 19.41$^*$ & 0.00$^*$ / 0.00$^*$ & 16 / 25 & 89.0$^*$ / 85.0$^*$ \\
Conditional look-back at the request step (StreamAff) & 30.14 / 28.46 & 32.85 / 31.45 & 33 / 55 & 76.5 / 72.0 \\
All observed frames at every step & 29.07 / 26.17$^*$ & 34.90$^*$ / 32.46 & 45 / 71 & 69.5$^*$ / 50.0$^*$ \\
\bottomrule
\end{tabular}
\par\smallskip
{\footnotesize $^*$: paired difference to StreamAff whose 95\% bootstrap interval over requests excludes zero (the counts are not tested).
Without the look-back, 11 / 5 History requests are still answered at the required step, but from the current frame, which is not a valid source.}
\end{table}

\paragraph{Can mask confidence determine when to respond?}
In the Future failure of \cref{fig:supp_pair}, the model produces a mask on the washing machine dial before the requested drawer appears.
A natural question is whether the predicted 3D mask confidence itself can decide when to respond, without an explicit wait option.
To test this, we fix the StreamAff model, the reads, and the pointer-selected source frame, and compare StreamAff's learned response score $p_{\mathrm{resp}} = 1 - p_{\varnothing}$ against $p_{\max}$, the maximum sigmoid probability across all 3D points of the selected frame after Zoom-and-Refine decoding (\cref{tab:supp_mask_conf}).
At the read level, $p_{\mathrm{resp}}$ cleanly separates required-response reads from wait reads (ROC-AUC 0.964 / 0.979, AP 0.869 / 0.825), whereas $p_{\max}$ lags substantially behind (ROC-AUC 0.787 / 0.783, AP 0.461 / 0.255) and suffers a 3.9--7.1$\times$ higher false-positive rate at 95\% recall (0.592 / 0.557 vs.\ 0.153 / 0.078).
Sweeping all distinct $p_{\max}$ values for end-to-end streaming triggering confirms this gap: even at its oracle-best StreamIoU threshold ($\tau = 0.957 / 0.939$), $p_{\max}$ trails StreamAff by 5.72 / 5.73 points (24.43 / 22.73) while almost never waiting (3.5\% / 0.5\% correct-wait rate and only 3.86 / 1.33 Future StreamIoU), barely improving over answering on any decoded mask (23.46 / 22.37).
Raising the threshold to match StreamAff's correct-wait rate (76.5\% / 72.0\%, at $\tau = 0.99925 / 0.99977$) collapses StreamIoU to 11.84 / 6.77, losing 18.31 / 21.69 points overall and degrading every answerable setting.
Because mask supervision ($\mathcal{L}_{\mathrm{grd}}$ in Eq.~(7) of the main paper) applies only when a valid target is present ($g=0$ on wait targets), a segmentation head trained solely for positive grounding remains overconfident on visually similar distractors when the part is absent, showing that mask confidence alone cannot replace explicit wait supervision.
Moreover, even if negative mask supervision were added to the 3D decoder, gating on $p_{\max}$ would still require running answer generation and Zoom-and-Refine 3D decoding at every waiting step (adding 0.76\,s per step), whereas placing wait supervision in the lightweight pointing head decides before decoding and avoids this cost while waiting.

\begin{table}[t]
\centering
\caption{\textbf{Learned response score vs.\ predicted 3D mask confidence} (SceneFun3D 15\,s / 30\,s).
Both signals use the same StreamAff model, reads, and pointer-selected source frame; $p_{\max}$ is the maximum sigmoid probability across all 3D points of the selected frame after Zoom-and-Refine decoding.
$^*$ marks a paired difference to StreamAff whose 95\% bootstrap interval over requests excludes zero.}
\label{tab:supp_mask_conf}
\small
\begin{tabular}{@{}lccc@{}}
\toprule
Decision signal & ROC-AUC $\uparrow$ & AP $\uparrow$ & FPR@95TPR $\downarrow$ \\
\midrule
Learned response score $p_{\mathrm{resp}} = 1 - p_{\varnothing}$ (StreamAff) & \textbf{0.964} / \textbf{0.979} & \textbf{0.869} / \textbf{0.825} & \textbf{0.153} / \textbf{0.078} \\
Mask confidence on selected frame $p_{\max}$ & 0.787$^*$ / 0.783$^*$ & 0.461$^*$ / 0.255$^*$ & 0.592 / 0.557 \\
\bottomrule
\end{tabular}

\vspace{2.5mm}
\begin{tabular}{@{}lccccc@{}}
\toprule
& \multicolumn{4}{c}{StreamIoU $\uparrow$} & Wait $\uparrow$ \\
\cmidrule(lr){2-5} \cmidrule(l){6-6}
Triggering rule & Avg. & Cur. & Hist. & Fut. & Unobs. \\
\midrule
StreamAff ($p_{\mathrm{resp}} > 0.5$, default) & \textbf{30.14} / \textbf{28.46} & \textbf{34.28} / \textbf{36.59} & 32.85 / 31.45 & \textbf{23.30} / \textbf{17.33} & 76.5 / 72.0 \\
$p_{\max} \ge \tau$ (best StreamIoU; $\tau = 0.957 / 0.939$) & 24.43$^*$ / 22.73$^*$ & 33.54 / 32.40 & \textbf{35.89} / \textbf{34.45} & 3.86 / 1.33 & 3.5 / 0.5 \\
$p_{\max} \ge \tau$ (matched Wait; $\tau = 0.99925 / 0.99977$) & 11.84$^*$ / 6.77$^*$ & 13.67$^*$ / 7.95$^*$ & 12.03$^*$ / 6.35$^*$ & 9.80$^*$ / 5.99$^*$ & 76.5 / 72.0 \\
\bottomrule
\end{tabular}

\vspace{1.5mm}
\begin{minipage}{\linewidth}
\footnotesize
Top: read-level separation across all 650 requests per set (450 positive reads at the required response step $a$; 1{,}775 / 4{,}408 negative reads from Future pre-response steps and Unobserved streams; 1 / 6 reads with no decoded mask receive $p_{\max}=0$). Paired differences ($p_{\mathrm{resp}} - p_{\max}$, 10{,}000 request-level bootstrap resamples): ROC-AUC $+0.178\,[+0.156, +0.200]$ / $+0.196\,[+0.175, +0.219]$; AP $+0.408\,[+0.356, +0.458]$ / $+0.570\,[+0.516, +0.620]$.
Bottom: end-to-end streaming decision sweeping all distinct $p_{\max}$ values (1{,}463 / 2{,}218 thresholds). StreamAff minus $p_{\max}$ in overall StreamIoU: $+5.72\,[+3.22, +8.20]$ / $+5.73\,[+3.50, +8.07]$ at the best $p_{\max}$ threshold, and $+18.31\,[+15.48, +21.22]$ / $+21.69\,[+18.76, +24.63]$ at the matched correct-wait rate (Current $+20.61^*$ / $+28.64^*$, History $+20.82^*$ / $+25.10^*$, Future $+13.50^*$ / $+11.34^*$). Answering on any decoded mask ($\tau = 0$) yields 23.46 / 22.37 StreamIoU.
\end{minipage}
\end{table}

\paragraph{Outcomes and response-time tolerance.}
\cref{tab:supp_errors} assigns every answerable request to one outcome and records which read triggered the response.
\cref{tab:supp_tolerance} reports StreamIoU@$k$ (Sec.~5.4 of the main paper), which also credits a response up to $k$ steps after the required step and scores it with the mask that the method produces at that later step.

\begin{table}[t]
\centering\small
\caption{\textbf{Outcome of every answerable request.} Share of the answerable requests of each setting (\%) by outcome, SceneFun3D
15\,s / 30\,s. On time = response at the required step; valid / invalid source = the selected frame is / is not a valid source.}
\label{tab:supp_errors}
\setlength{\tabcolsep}{3pt}
\begin{tabular}{@{}l ccccc@{}}
\toprule
Setting & On time, valid source & On time, invalid source & Early & Late & No response \\
\midrule
Current & 84.7 / 84.7 & 6.0 / 8.7 & 0 / 0 & 4.7 / 4.0 & 4.7 / 2.7 \\
History & 69.3 / 73.3 & 20.7 / 18.0 & 0 / 0 & 2.0 / 3.3 & 8.0 / 5.3 \\
Future & 50.0 / 43.3 & 0 / 0 & 22.0 / 36.7 & 12.7 / 11.3 & 15.3 / 8.7 \\
All & 68.0 / 67.1 & 8.9 / 8.9 & 7.3 / 12.2 & 6.4 / 6.2 & 9.3 / 5.6 \\
\midrule
\multicolumn{6}{@{}l}{Which read triggered the response (\%, 15\,s / 30\,s)} \\
\midrule
\multicolumn{3}{@{}l}{On-time History, by the look-back at the request step} & \multicolumn{3}{l}{91.9 / 96.4} \\
\multicolumn{3}{@{}l}{On-time Future, by a later new frame} & \multicolumn{3}{l}{100 / 100} \\
\multicolumn{3}{@{}l}{On-time Current, by the current frame} & \multicolumn{3}{l}{84.6 / 82.1} \\
\multicolumn{3}{@{}l}{Early Future, by the look-back} & \multicolumn{3}{l}{60.6 / 67.3} \\
\multicolumn{3}{@{}l}{Unobserved answered (23.5 / 28.0\% of them), by the look-back} & \multicolumn{3}{l}{57.4 / 58.9} \\
\multicolumn{3}{@{}l}{History with an invalid source, by the look-back} & \multicolumn{3}{l}{64.5 / 81.5} \\
\bottomrule
\end{tabular}
\par\smallskip
{\footnotesize Late: a response after the required step. Most early Future answers happen at the request step itself (25 of 33, 46 of 55),
as do most Unobserved false answers (34/47, 38/56). The look-back brings almost all on-time History answers and is also the main source
of early Future answers, Unobserved false answers and wrong History frames.}
\end{table}

\begin{table}[t]
\centering\small
\caption{\textbf{StreamIoU under a response-time tolerance.} A response up to $k$ steps (seconds) after the labelled response step also
counts, scored with the real mask produced at that later step. $k{=}0$ is the StreamIoU of Tab.~2 in the main paper.}
\label{tab:supp_tolerance}
\setlength{\tabcolsep}{4pt}
\begin{tabular}{@{}l cccc cccc@{}}
\toprule
 & \multicolumn{4}{c}{15\,s streams} & \multicolumn{4}{c}{30\,s streams} \\
\cmidrule(lr){2-5}\cmidrule(lr){6-9}
Method & $k{=}0$ & $k{=}1$ & $k{=}2$ & $k{=}3$ & $k{=}0$ & $k{=}1$ & $k{=}2$ & $k{=}3$ \\
\midrule
Prompted Qwen3-VL-8B decider & 19.25 & 21.10 & 21.55 & 22.02 & 19.02 & 20.25 & 20.86 & 21.08 \\
Prompted Qwen3-VL-4B decider & 21.43 & 22.99 & 23.31 & 23.50 & 19.14 & 19.79 & 20.10 & 20.10 \\
StreamAff           & \textbf{30.14} & \textbf{31.41} & \textbf{31.79} & \textbf{32.13} & \textbf{28.46} & \textbf{28.97} & \textbf{29.02} & \textbf{29.39} \\
\bottomrule
\end{tabular}
\par\smallskip
{\footnotesize Paired gaps, StreamAff minus the prompted Qwen3-VL-8B decider, $k{=}1/2/3$: 15\,s +10.31 [7.55, 13.20], +10.24 [7.41, 13.19], +10.12 [7.28, 13.00];
30\,s +8.72 [6.17, 11.25], +8.15 [5.60, 10.65], +8.31 [5.77, 10.84] (95\% paired bootstrap over requests).
StreamAff minus the prompted Qwen3-VL-4B decider: 15\,s +8.42 [5.64, 11.31], +8.48 [5.69, 11.31], +8.63 [5.81, 11.56]; 30\,s +9.18 [6.54, 11.86], +8.92 [6.28, 11.55], +9.29 [6.64, 11.88].
StreamAff thus stays at least 8.15 points ahead of both prompted deciders for $k{=}0$ to $3$.
At $k{=}0$, with scenes resampled instead of requests, the difference and its scene-level 95\% interval are +10.89 [6.73, 15.12] on 15\,s and
+9.44 [6.34, 13.13] on 30\,s streams to the prompted Qwen3-VL-8B decider, and +8.72 [4.80, 12.95] and +9.32 [4.96, 13.94] to the prompted
Qwen3-VL-4B decider. With any delay allowed, every late response
being scored in the same way: StreamAff 32.13 / 29.39, prompted Qwen3-VL-8B 22.18 / 21.26, prompted Qwen3-VL-4B 23.63 / 20.10.
StreamAff is rarely late (15\,s: 16 / 4 / 9 responses at 1 / 2 / $\geq$3 steps late);
most of its gain is on Future requests (15\,s: 23.30 $\to$ 28.28 at $k{=}3$).}
\end{table}

\FloatBarrier

\section{Selecting the Frame and Segmenting the Part}
\label{sec:supp_where}

On 15\,s streams StreamAff reaches a StreamIoU of 30.14.
With the response placed at the required step and the source frame selected by the model, its mIoU is 37.25; with a valid source frame given as well, the mIoU averaged over the valid frames is 40.18 (\cref{tab:supp_decomp}).

\begin{table}[t]
\centering\small
\caption{\textbf{From the streaming score to the mask on a given frame.} Four conditions on the same model (SceneFun3D 15\,s / 30\,s).
The first row is StreamIoU. The other rows are mIoU with the response placed at the required step, so that timing plays no role; a
selected frame that is not a valid source scores 0. In the last two rows the mask is decoded, with the same generated part phrase, on
every valid source frame of the request.}
\label{tab:supp_decomp}
\setlength{\tabcolsep}{5pt}
\begin{tabular}{@{}l l c c@{}}
\toprule
Condition & Metric & Avg. & History \\
\midrule
Streaming: StreamAff decides when to respond and selects the frame & StreamIoU & 30.14 / 28.46 & 32.85 / 31.45 \\
Response at the required step, StreamAff selects the frame & mIoU & 37.25 / 35.98 & 34.53 / 33.69 \\
Response at the required step, valid source frame given (mean over the valid frames) & mIoU & 40.18 / 38.74 & 42.81 / 40.00 \\
Response at the required step, valid source frame given (best of the valid frames) & mIoU & 43.76 / 43.29 & 47.80 / 47.19 \\
\bottomrule
\end{tabular}
\end{table}

The masks of StreamAff are defined on the 3D points of the selected source frame: they are not fused across frames and inference uses no camera pose, so the selected frame enters the score directly.
On History requests of 15\,s streams, a given valid source frame raises the mIoU from 34.53 to 42.81, so the choice of the frame still matters after a response on time.
Among the responses at the required step, the selected frame is not a valid source in 11.56\% / 11.70\% of cases, lower than both prompted deciders on 30\,s streams (12.33\% for Qwen3-VL-8B and 15.45\% for Qwen3-VL-4B) and comparable on 15\,s streams (10.92\% and 9.34\%) despite StreamAff answering substantially more requests on time (76.9\% vs.\ 52.9\% and 57.1\%, Sec.~5.1 of the main paper).

\paragraph{Zoom and Refine.}
Removing only the Zoom branch and retraining the model lowers StreamIoU by 5.38 / 2.83 points, and dropping the Refine head at test time lowers it by 4.16 / 3.91 points; all four differences are significant.

\paragraph{Agreement of the answer and the mask.}
A response consists of an answer in language and a mask, and we check whether the two refer to the same part.
To separate text--mask consistency from timing and localization failures, the check covers the 261 requests (140 / 121) that StreamAff answers at the required step, from a valid source frame and with an IoU of at least 0.5.
The judge is Qwen3-VL-30B.
It sees the source frame with the outline of the predicted mask, the request and the generated answer, and decides whether the parent object and the part named in the answer are the outlined ones.
The parent object agrees in 91.2\% of the cases, the part in 85.4\%, and both in 85.1\%.

\paragraph{Failure cases.}
\cref{fig:supp_boundary} shows three Current requests at the limit of the observation conditions.
All three responses are on time.
In the first and third rows, StreamAff selects a valid frame but misses the target part.
In the middle row, too little of the drawer is visible for the selected frame to qualify as a valid source.

{\par\medskip\noindent
\begin{minipage}{\linewidth}
\centering
\includegraphics[width=\linewidth,height=0.35\textheight,keepaspectratio]{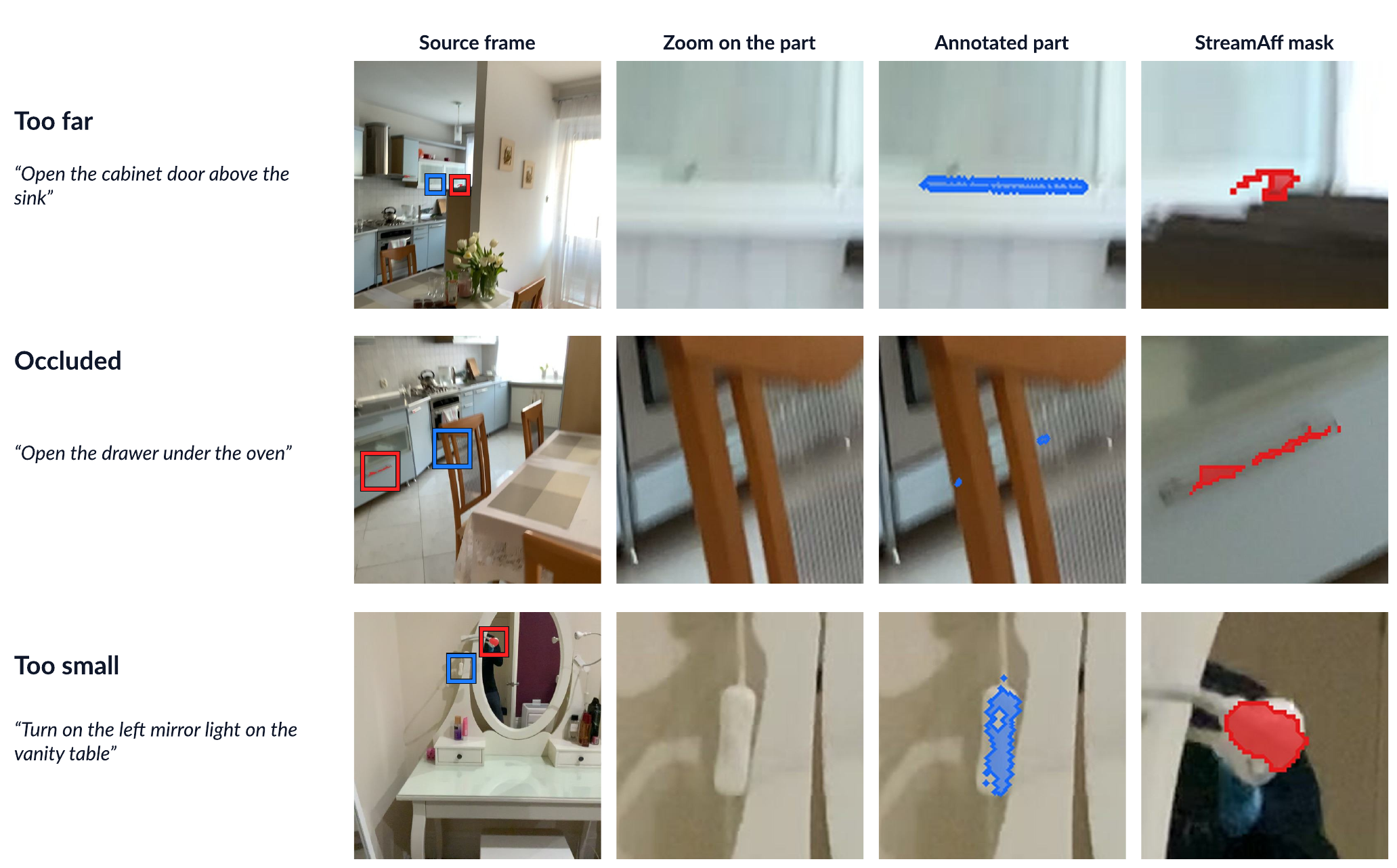}
\captionof{figure}{\textbf{Boundary cases of the observation conditions} (SceneFun3D 15\,s, Current requests). In all three cases StreamAff responds
at the required step, but its mask does not land on the part. \emph{Too far:} the handle is 4.8\,m away and covers about 200 pixels; the
selected frame is a valid source, yet the mask falls on a nearby spot (IoU 0). \emph{Occluded:} StreamAff grounds the drawer on the frame
just before the request, where a chair hides about 90\% of it. About 10\% of the part is visible, below the 50\% threshold of a valid
source frame (\cref{sec:supp_labelling}), so the response scores 0; whether a whole stream is excluded is decided separately, by the depth
criterion of \cref{sec:supp_sampling}. \emph{Too small:} the switch on the mirror arm covers about 100 pixels; the mask falls on the top of
the mirror frame (IoU 0). Blue box and fill: the annotated part (the benchmark's visible part on the source frame). Red box and fill:
StreamAff's mask, which lies outside the blue box in all three cases; the last column zooms on the mask at the same scale.}
\label{fig:supp_boundary}
\end{minipage}\par\medskip}

\FloatBarrier

\section{Benchmark Construction and Data Splits}
\label{sec:supp_benchmark}

\subsection{Evidence Labelling}
\label{sec:supp_labelling}

As described in Sec.~3.2 of the main paper, the part annotations of SceneFun3D~\cite{delitzas2024scenefun3d} and of ScanNet++~\cite{yeshwanth2023scannetpp} (through Articulate3D~\cite{halacheva2024articulate3d}) are projected into every frame with the camera pose and the depth map, which yields a mask on the 3D points of that frame.
To ensure that a \emph{valid} source frame provides a clearly observable and geometrically accurate 3D target mask rather than a barely visible glimpse or a noisy projection, SceneFun3D requires three checks with the same thresholds for both stream lengths:
(i) visibility (rejecting heavily occluded views): at least 50\% of the annotated part is visible, where a projected reference point counts as visible if its depth agrees with the depth map within 5\,cm in a $3\times3$ neighbourhood to tolerate depth and pose jitter;
(ii) size (rejecting tiny projections): the visible reference points cover at least 100 pixels;
(iii) geometric consistency (filtering depth-edge and projection artifacts): at least 80\% of the visible reference points lie within 2\,cm of a frame-mask point, and vice versa.
To guard against single-sensor depth artifacts, the visibility check must also hold on an independent depth sensor.
Because the 3D mask is grounded on the target part alone while a mentioned reference object may appear in other frames, all three checks apply only to the target part.
A frame is \emph{absent} if the part is verified to be out of view, and \emph{unlabelled} otherwise.

On ScanNet++, evidence labels are derived directly from the scene mesh.
To avoid treating isolated noisy voxels as valid observations while keeping evaluation targets clearly observable, a part is marked visible if at least 3 of its annotated 2\,cm voxels are seen, and an evaluation source frame requires at least 6.
To ensure that \emph{absent} labels are conservative, a frame is marked absent only if the part lies outside the view frustum with a 20-pixel safety margin (excluding ambiguous image-border cases) or if mesh ray casting hits no cell of the part (confirming occlusion).
As a further safeguard against mesh-discretization artifacts that could unfairly penalize an early response, a request is dropped if any frame labelled absent before the response still exposes 20 or more pixels of the part at full resolution.
Camera poses are used for annotation only.

\subsection{Response Steps and Sampling}
\label{sec:supp_sampling}

The required response step $a$ and the valid sources $S$ follow from the evidence timeline without manual timing labels (Sec.~3.2 of the main paper).
On SceneFun3D, requests are placed under three constraints: at least 5 frames are observed before the request step, the request step lies more than 2 steps before the end of the stream, and a Future request has to wait at least 2 steps.
These constraints give the ranges in Tab.~1 of the main paper.
Unobserved requests require all frames of the stream to be absent, and a request whose setting depends on an unlabelled frame is discarded and never used as a negative.
A stream in which the target part is confirmed to be occluded in any frame is excluded.
An occlusion is confirmed if at least 80\% of the part points that the rendering of the laser scan shows as visible lie behind the measured depth by more than 10\,cm or three times the median depth error, whichever is larger, \ie, if a foreground object that is absent from the scan hides the part; this is the case for 4{,}085 of the 497{,}487 checked frame--task pairs.
On ScanNet++, the request step lies between 20\% and 80\% of the stream length, a History request needs a visible frame at least 2 steps before the request, and a Future request needs a visible frame as the first non-absent frame after the request, again at least 2 steps later.

The 28 scenes of the 30\,s SceneFun3D evaluation set are among the 30 scenes of the 15\,s set, and none of the 30 SceneFun3D evaluation visits appear in the training split of Ground3D-LMM~\cite{ground3dlmm}.
The 892 ScanNet++ training clips (77 scenes) share no scene with the 56 ScanNet++ evaluation scenes.

For the 15\,s evaluation set, the labelling and sampling rules yield 1{,}175 candidate windows on the 30 evaluation scenes.
The review of the requests and frames was run on 1{,}032 requests (780 answerable, 252 Unobserved) and excluded 145 of them, and the final set has 650 requests (225 Easy, 225 Hard, 200 Unobserved).
For the 30\,s set, the review excluded 706 of 3{,}682 windows (652 without accepted visual evidence of the part, 54 with uncertain evidence), and 650 were selected.

\paragraph{Easy and Hard.}
An answerable request is Hard if the smallest projected area of the part over its valid source frames is below 200 pixels, or if instance disambiguation is required, and Easy otherwise.
On SceneFun3D, instance disambiguation means that a similar part competes in the frames observed by the response step, so that position, direction or order is needed to select the target; this was verified visually during benchmark construction.
On ScanNet++, it means that a part of the same type is also asked about in the scene, or that the request contains an ordinal or a direction word; all Hard requests of ScanNet++ are of this kind.

\subsection{Requests and Reference Answers}
\label{sec:supp_qa}

\paragraph{SceneFun3D.}
The requests are the task descriptions of SceneFun3D.
The reference answers are drafted by Qwen3-VL-8B and verified by Qwen3-VL-30B, both of which see the source frame with the outline of the annotated part.
A later pass re-checks every answer for the identity of the part and of its parent object, the spatial reference, the coverage of the annotated mask, whether the action is supported, and whether the answer is direct.
Samples that remained undetermined after these passes are discarded.

\paragraph{ScanNet++.}
Request--answer pairs are rewritten from the part descriptions of OpenTarget~\cite{huang2025openground} and checked for target consistency and disambiguation (Sec.~3.2 of the main paper).
A rewritten request keeps only facts of the target part's description and adds the minimum needed to distinguish it from sibling parts.
A part is dropped if its description cannot be told apart from that of a sibling, or if the ordinal or direction it would need is not given; the order of stacked drawers and switches is taken from the 3D boxes, and every attribute is checked against an overlay of the mask on a source frame.
Of 226 candidate parts, 146 are kept and 80 dropped: 9 fail the overlay check and 71 are ambiguous or contradict the label or the geometry.

\paragraph{Final review.}
As stated in the main paper, all evaluation samples undergo a final manual review: after the model-assisted checks above, the authors checked all requests and reference answers by hand.
In this review the authors also checked, on the frame at the labelled response step of every Current and Future request, that the target part can be identified.

\subsection{Check of the Labelled Response Steps}
\label{sec:supp_labelcheck}

An early response is penalized only if the part was indeed not visible before the labelled response step.
\cref{tab:supp_labels} checks the labels of every SceneFun3D evaluation request against the per-frame visibility of its target part.

\begin{table}[t]
\centering\small
\caption{\textbf{Response-step labels checked against per-frame visibility.} For every request of the SceneFun3D evaluation sets
(150 per answerable setting, 200 Unobserved per stream length), the visibility state of the target part at the labelled response step.}
\label{tab:supp_labels}
\setlength{\tabcolsep}{4pt}
\begin{tabular}{@{}l cc@{}}
\toprule
 & 15\,s streams & 30\,s streams \\
\midrule
Current: part visible at the response step & 150 / 150 & 150 / 150 \\
\quad target pixels (p10 / p50 / p90) & 123 / 244 / 666 & 128 / 240 / 602 \\
History: part out of view at the request & 150 / 150 & 150 / 150 \\
\quad seconds since last visible (p10 / p50 / p90) & 1 / 5 / 10 & 1 / 5 / 22 \\
Future: part visible at the response step & 150 / 150 & 150 / 150 \\
\quad out of view one step before & 150 / 150 & 150 / 150 \\
\quad target pixels at the response step (p10 / p50 / p90) & 124 / 239 / 703 & 120 / 237 / 581 \\
Unobserved: target pixels on any frame & 0 on all 200 & 0 on all 200 \\
\bottomrule
\end{tabular}
\par\smallskip
{\footnotesize Future, frames from the request up to the step before the response step (labelled ``wait''): 0 target pixels or no stored evidence on all
479 (15\,s) and 748 (30\,s) frames. Waiting time from request to response step: 2--8 s (15\,s), 2--19 s (30\,s).}
\end{table}

\FloatBarrier

\section{Implementation and Baseline Protocols}
\label{sec:supp_impl}

\subsection{Model and Trainable Modules}
\label{sec:supp_modules}

StreamAff is initialized from Ground3D-LMM-4B-Joint~\cite{ground3dlmm}, built on Qwen3-VL-4B~\cite{bai2025qwen3vl}, with both the visual and 3D encoders kept frozen.
Each incoming RGB-D frame is downsampled to $640\times480$ pixels ($480\times640$ in portrait) and encoded on arrival into $N=176$ 3D-aware visual tokens without separate point tokens: we resize the image to $512\times352$ ($352\times512$ in portrait) to obtain a $16\times11$ ($11\times16$) token grid, pool the 3D voxel features onto these tokens via depth-based pixel--voxel correspondence, and add them as a residual through a projection and an adapter.
We also run the visual encoder at the $640\times480$ resolution ($20\times15$ or $15\times20$ tokens) to supply high-resolution RGB features for local voxel fusion and the Zoom branch.

In total, StreamAff trains 25.0\,M parameters: rank-16 LoRA weights on the attention projections of all 36 LLM layers (11.80\,M), the 3D mask decoder (7.13\,M), the voxel-to-token projection and adapter (0.72\,M and 2.63\,M), local RGB-to-voxel fusion (0.09\,M), the Zoom branch (0.74\,M), the Refine head (0.19\,M), and the pointing head with the wait option (1.72\,M).
At response time, the Zoom branch scores 48 fine windows ($6\times8$ grid) and 4 coarse windows ($2\times2$ grid) on the selected frame, then re-encodes the top two fine and top two coarse crops with the long side at 768 pixels.
These four crops yield 1{,}872--2{,}016 visual tokens per request, which are pooled onto the 3D voxels for mask decoding rather than fed into the language model.

\subsection{Pointing Head and Wait Option}
\label{sec:supp_head}

For a read $\mathcal{R}$, the pointing head assigns a score $s_{ij}$ to patch $j$ of frame $i\in\mathcal{R}$ and a score $s_{\varnothing}$ to the wait option, and normalizes all of them together:
\begin{equation}
\label{eq:supp_softmax}
p_{ij}=\frac{\exp(s_{ij})}{Z},\qquad
p_{\varnothing}=\frac{\exp(s_{\varnothing})}{Z},\qquad
Z=\exp(s_{\varnothing})+\sum_{i'\in\mathcal{R}}\sum_{j'=1}^{N}\exp(s_{i'j'}).
\end{equation}
Eq.~(3) of the main paper follows, $p_{\mathrm{resp}}=\sum_{i\in\mathcal{R}}\sum_{j}p_{ij}=1-p_{\varnothing}$, and the source frame is $i^\star=\argmax_{i\in\mathcal{R}}\max_j p_{ij}$.

\paragraph{Scores.}
The request is placed after all frames of the read, and the query $\mathbf{h}$ is the last-layer hidden state at the last input token.
With $\mathbf{H}_{ij}$ the last-layer hidden state of patch $j$ of frame $i$, the head computes $\mathbf{h}_q=W_h\,\mathrm{LN}(\mathbf{h})$ and $\mathbf{v}_{ij}=W_v\,\mathrm{LN}(\mathbf{H}_{ij})$, both of dimension 256, and
\begin{equation}
s_{ij}=\mathrm{MLP}\bigl([\mathbf{h}_q;\,\mathbf{v}_{ij};\,\mathbf{h}_q\odot\mathbf{v}_{ij}]\bigr),\qquad
s_{\varnothing}=\mathrm{MLP}_{\varnothing}\bigl([\mathbf{h}_q;\,\mathbf{e}_K;\,\mathbf{e}_w]\bigr),
\end{equation}
where both MLPs have one hidden layer of 256 units with GELU, $\mathbf{e}_K$ embeds the number of frames in the read, and $\mathbf{e}_w$ embeds the number of steps since the request (both capped at 16).
The scores enter the softmax of \cref{eq:supp_softmax} directly, without a temperature.
The agent waits if $p_{\varnothing}\ge0.5$ and otherwise responds with the frame of the highest-scoring patch.

\subsection{Grounding Query}
\label{sec:supp_query}

As in the main paper, the grounding query $\mathbf{z}_{\mathrm{part}}$ is the mean hidden state of the part phrase in the generated answer, which is bounded by a pair of delimiter tokens and followed by a \texttt{<SEG>} marker (Fig.~3 of the main paper).
The \texttt{<SEG>} token serves solely as a decoding trigger signalling that the grounded phrase is complete, at which point the decoder $\mathcal{G}$ predicts the 3D mask on the selected frame from $\mathbf{z}_{\mathrm{part}}$.

\subsection{Training}
\label{sec:supp_training}

All trainable parts are optimized jointly with the loss of Eq.~(7) of the main paper.
Both the full model and all ablation variants use the same training data, order, and random seed, are trained for 5{,}040 updates, and are evaluated at the final checkpoint.

\paragraph{Optimization.}
We train on four H200 GPUs in bf16 with AdamW (weight decay 0.05, gradient clipping at norm 10), a learning rate of $10^{-5}$ for the LoRA weights and $10^{-4}$ for all other trained parts, no warm-up, and a polynomial decay with power 0.95 that is stepped every 360 updates.
LoRA has rank 16 ($\alpha=32$, dropout 0.1) on the query, key, value and output projections of all 36 layers of the language model.

\paragraph{Batches.}
An update is either a 15\,s or a 30\,s update, in the ratio 3\,:\,1.
Each GPU processes two grounding clips and one decision event per update, \ie, eight clips and four events in total; one clip and one event of every update come from ScanNet++ (892 training clips), the others from SceneFun3D (2{,}400 clips of 15\,s and 800 of 30\,s).
Over the 5{,}040 updates the model thus sees 20{,}160 decision events: 11{,}340 from 15\,s SceneFun3D streams, 3{,}780 from 30\,s SceneFun3D streams and 5{,}040 from ScanNet++; 14{,}346 carry a wait target and 5{,}814 a response target.

\paragraph{Reads at the request step.}
During training, the read at the request step contains only the current frame with probability 0.5 and all observed frames otherwise (at most 8{,}192 context tokens).
If the read contains only the current frame and that frame is not a valid source, the target is to wait.
After the request step every read contains only the new frame.
At inference, the look-back reads all observed frames in chunks of at most 32 frames and carries the four frames with the highest patch scores from one chunk to the next, so a longer history is scanned completely and is not truncated.
With chunks of 16 frames the StreamIoU on 30\,s streams is 28.39 instead of 28.46, a difference that is not significant; a 15\,s stream fits into one chunk in both cases.

\paragraph{Latency.}
On a single H200 GPU, StreamAff processes each incoming frame in 0.25\,s (median) before the request and 0.31\,s afterwards on 15\,s streams, with answer generation and Zoom-and-Refine 3D mask decoding adding 0.76\,s upon response (measured on 30 requests, Sec.~5.4 of the main paper).
Across all 455 responses to the 650 requests of the 15\,s evaluation set, a full response step takes 0.96\,s on average and 1.16\,s at the 95th percentile.
Because answer generation and 3D mask decoding are invoked only once upon triggering a response, they can be executed asynchronously without blocking the lightweight 0.25--0.31\,s per-frame ingestion and decision loop.

\paragraph{Loss weights.}
The mask loss is the sum of the decoder loss (binary cross-entropy plus Dice with weight 4), the Zoom region loss with weight 0.25, and the Refine loss with weight 1.

\subsection{External Deciders}
\label{sec:supp_decider_interface}

To isolate timing and frame selection from language and 3D mask generation, the external decider only decides when to respond and which frame to use, while StreamAff generates the answer, part phrase, and 3D mask on that chosen frame; external-decider baselines thus differ from StreamAff solely in response timing and source-frame selection.

At the request step and each subsequent step $t$, the decider receives the task request along with all RGB frames observed so far ($0\dots t$, up to 15 or 30 images in chronological order), each prefixed by ``Frame $k$:'', and decodes greedily using the prompt:
\begin{quote}\small
\emph{These are frames 0 to $t$ of a video recorded while walking through a room, in time order. Task: ``\{task\}''. Which object part would a person have to interact with to do this task? If that part is clearly visible in at least one frame, answer with only the number of the frame where it is seen best. If it is not visible in any frame, answer with only NONE.}
\end{quote}
If the output contains ``none'' or lacks an integer index within $[0, t]$, the agent waits; otherwise, the first step returning a valid index triggers a response on that frame.
The same prompt and parsing rule apply to both prompted deciders across both stream lengths.

\subsection{Offline Baselines}
\label{sec:supp_offline}

The offline methods of Tabs.~2 and~3 of the main paper are run with their released code and weights; none of them is trained or fine-tuned by us.
Each method receives the frames of the complete stream, and its output is scored on the 3D points of the annotated valid source frames, as the mean IoU over these frames.

\paragraph{Request-prefix variant.}
For Fun3DU~\cite{corsetti2025fun3du} and T-FunS3D~\cite{feng2026tfuns3d}, a cell of Tab.~2 also reports a StreamIoU for a request-prefix variant, run on the reduced frames of \cref{sec:supp_modules} and scored as an immediate response at the request step.
For Current and History requests, the method receives only the frames observed up to the request step; for Future requests, an immediate response is always early and scores 0.
The mIoU in the same cell uses the complete stream.

\paragraph{Fun3DU and T-FunS3D.}
Both methods expect a reconstructed scene.
We fuse the RGB-D frames of the stream into a point cloud with the camera poses, let the method vote over the views on this cloud (8\,mm voxels), and map the result back to the points of each frame.
Fun3DU uses its official language model (Llama~3.1 8B, called as in the official pipeline) and T-FunS3D uses Qwen3-14B; the other components are the frozen models of the official pipelines.

\paragraph{AffordBot.}
AffordBot~\cite{wang2025affordbot} uses the officially released full-scene Mask3D functional-part proposals and Qwen2.5-VL-72B in 8-bit precision.
Its scene-level mask is defined on the laser scan and is transferred to the points of a frame through nearest neighbours within 2\,cm.

\paragraph{Sa2VA, GLOVER++ and AffordanceVLM.}
Sa2VA-8B~\cite{yuan2025sa2va}, GLOVER++~\cite{ma2025gloverpp} and AffordanceVLM~\cite{wu2025ragnet} are used with their official weights, zero-shot.
Each predicts a 2D mask per frame, which is lifted to the 3D points of that frame through its valid depth pixels.
GLOVER++ and AffordanceVLM process each frame independently, and Sa2VA processes the stream with its official video inference.

\subsection{Limitations}
\label{sec:supp_limitations}

After the request step, StreamAff reads each newly arriving frame alone to keep per-step latency constant.
As a result, it cannot combine a reference object seen before the request with a target part that appears only later; resolving such cross-frame references would require carrying a lightweight visual state across post-request reads.

In addition, both training and evaluation focus on 15\,s and 30\,s streams.
To test zero-shot length extrapolation beyond the 30\,s training horizon, we prepended 60 earlier frames from the same visit to the 15\,s streams (75\,s in total).
On requests whose labels remained unchanged, Future StreamIoU dropped by 6.66 points (42 requests) and the Unobserved correct-wait rate dropped by 14.8 percentage points (27 requests).
Because post-request frames are read alone, the prepended frames enter only through the request-step look-back, which must screen a much larger pool of distractor frames than encountered during training.
Extending StreamAff to multi-minute streams will therefore require training on longer episodes or bounding the look-back window.

\FloatBarrier

\end{document}